\documentclass[12pt,preprint,review]{elsarticle}

\usepackage[utf8]{inputenc}
\usepackage{amsmath,amssymb}
\usepackage{verbatim}
\usepackage{soul,color}
\usepackage{bbold}
\usepackage{adjustbox}
\usepackage{subcaption}
\usepackage{graphicx}
\usepackage{lipsum}
\usepackage{multirow}
\usepackage{booktabs}
\usepackage{mathtools}
\usepackage{balance}
\usepackage{placeins}
\usepackage{hyperref}
\usepackage{bm}
\usepackage{algpseudocode}
\usepackage{algorithm}

\newcommand{\bt}{\fontseries{b}\selectfont}
\newcommand{\rpm}{\raisebox{.2ex}{$\scriptstyle\pm$}}
\usepackage{mfirstuc} 
\newcommand{\addReviewer}[2]{
  \expandafter\newcommand\csname #1\endcsname[1]{{\bf \color{#2} \capitalisewords{#1}:\,##1}}
  \expandafter\newcommand\csname #1cor\endcsname[2]{{\color{#2} \capitalisewords{#1}:\,\st{##1}{\bf ##2}}}
  \expandafter\newcommand\csname #1color\endcsname{#2}
}

\DeclarePairedDelimiterX{\infdivx}[2]{}{}{%
  #1\;\delimsize|\delimsize|\;#2%
}

\definecolor{asparagus}{rgb}{0.53, 0.66, 0.42}
\definecolor{alizarin}{rgb}{0.82, 0.1, 0.26}
\addReviewer{jary}{asparagus}
\addReviewer{simone}{alizarin}

\journal{Neurocomputing}

\begin{document}

\begin{frontmatter}

\title{Bayesian Neural Networks With Maximum Mean Discrepancy Regularization}

\author[sapienza]{Jary Pomponi}
\ead{jary.pomponi@uniroma1.it}

\author[sapienza]{Simone Scardapane}
\ead{simone.scardapane@uniroma1.it}

\author[sapienza]{Aurelio Uncini}

\address[sapienza]{Department of Information Engineering, Electronics and Telecommunications (DIET), Sapienza University of Rome, Italy}




\begin{abstract}
    Bayesian Neural Networks (BNNs) are trained to optimize an entire distribution over their weights instead of a single set, having significant advantages in terms of, e.g., interpretability, multi-task learning, and calibration. Because of the intractability of the resulting optimization problem, most BNNs are either sampled through Monte Carlo methods, or trained by minimizing a suitable Evidence Lower BOund (ELBO) on a variational approximation. In this paper, we propose an optimized version of the latter, wherein we replace the Kullback-Leibler divergence in the ELBO term with a Maximum Mean Discrepancy (MMD) estimator, inspired by recent work in variational inference. After motivating our proposal based on the properties of the MMD term, we proceed to show a number of empirical advantages of the proposed formulation over the state-of-the-art. In particular, our BNNs achieve higher accuracy on multiple benchmarks, including several image classification tasks. In addition, they are more robust to the selection of a prior over the weights, and they are better calibrated. As a second contribution, we provide a new formulation for estimating the uncertainty on a given prediction, showing it performs in a more robust fashion against adversarial attacks and the injection of noise over their inputs, compared to more classical criteria such as the differential entropy. 
\end{abstract}

 
\end{frontmatter}

\section{Introduction}
Deep Neural Networks (DNNs) are currently the most widely used and studied models in the machine learning field, due to the large number of problems that can be solved very well with these architectures, such as image classification \cite{he2015delving}, speech processing \cite{rethage2018wavenet}, image generation \cite{li2017mmd}, and several others. Despite their empirical success, however, these models have a number of open research problems. Among them, how to quantify the uncertainty of an individual prediction remains challenging \cite{kendall2017uncertainties}. A concrete measure of uncertainty is critical for several real-world applications, e.g., driverless vehicles, out-of-distribution detection, and medical applications \cite{kwon2020uncertainty}. 


The principal approach to model the uncertainty in these models is based on Bayesian statistics. Bayesian Neural Networks (BNNs) model the parameters of a DNN as a probability distribution computed via the application of the Bayes' rule, instead of a single fixed point in the space of parameters \cite{mackay1992practical}. Despite a wealth of theoretical and applied research, the challenge with these BNNs is that codifying a distribution over the weights remain difficult, mainly because: (1) the minimization problem is intractable in the general case, and (2) we need to specify prior knowledge (in the form of a prior distribution) over the parameters of the network, and in most datasets the results can be sensible to this choice \cite{wenzel2020good}. On top of this, applying these principles to a Convolutional Neural Network (CNN) is even harder, because of the depth of these networks in practice.

In the last years, different approaches have been proposed to build and/or train a BNN, outlined later in Section \ref{sec:related_work}. In general, these approaches can either avoid the minimization problem altogether and sample from the posterior distribution \cite{chen2015convergence}, or they can solve the optimization problem in a restricted class of variational approximations \cite{blundell2015weight}. The latter approach (referred to as Variational Inference, VI) has become extremely popular over the last years thanks to the possibility of straightforwardly leveraging automatic differentiation routines common in deep learning frameworks \cite{blei2017variational}, and avoiding a large quantity of sampling operations  during the inference phase. However, the empirical results of BNNs remain sub-optimal in practice \cite{wenzel2020good}, and ample margins exist to further increase their accuracy, robustness to the choice of the prior distribution, and calibration of the classification models.

\subsection{Contributions of the work}
\label{subsec:contributions}
In this paper, we partially address the aforementioned problems with two innovations related to BNNs. Firstly, we propose a modification of the commonly used optimization procedure in VI for BNNs. In particular, we leverage across recent works in variational auto-encoding \cite{2017arXiv170602262Z} to propose a modification of the standard Evidence Lower BOund (ELBO) minimized during the BNN training. In the proposed approach, we replace the Kullback-Leibler term on the variational approximation with a more flexible Maximum Mean Discrepancy (MMD) estimator \cite{Gretton:2012:KTT:2188385.2188410}. After motivating our proposal, we perform an extensive empirical evaluation showing that the proposed BNN can significantly improve over the state-of-the-art in terms of classification accuracy, calibration, and robustness in the selection of the prior distribution. Secondly, we provide a new definition to measure the uncertainty on the prediction over a single point. Different from previous state-of-the-art approaches \cite{kwon2020uncertainty}, our formulation provides a single scalar measure also in the multi-class case. In the experimental evaluation, we show that it performs better when defending from an adversarial attacks against the BNN using a simple thresholding mechanism.

\subsection{Related work}
\label{sec:related_work}

\subsubsection{VI training for BNNs}
The idea to apply Bayesian methods on neural networks has been studied widely during the years. In \cite{buntine1991bayesian} the authors were the first to propose several Bayesian methods applied to the networks, but only in \cite{hinton1993keeping} the first VI method was proposed as a regularization approach. In \cite{mackay2003information} and \cite{neal2012bayesian} posterior's approximations were investigated, in the first case using a Laplacian approximation and in the second one by a Monte Carlo approach. Only recently, the first practical VI training techniques were advanced in \cite{graves2011practical}. In \cite{blundell2015weight}, this approach was extended and an unbiased way of updating the posterior was found. Dropout has also been proposed as an approximation of VI \cite{gal2015dropout,kingma2015variational}. 

While these methods can be applied in general for most DNNs, few works were carried out in the context of image classification, due to the complexity and the depth of the networks involved in these tasks, combined with the inner difficulties of VI methods. In \cite{gal2015bayesian} and \cite{2019arXiv190102731S} the authors used Bayesian methods to train CNNs, while in \cite{kendall2017uncertainties} and \cite{kwon2020uncertainty} the authors proposed two alternatives that work also for CNNs, to measure the uncertainty of a classification, using the posterior distribution. 

Almost all the works devoted to VI training of BNNs have considered the standard ELBO formulation \cite{blei2017variational}, where we minimize the sum of a likelihood term and the Kullback-Leibler (KL) divergence with respect to the variational approximation. However, recently several works have put forward alternative formulations of the ELBO term replacing the KL term with separate divergences \cite{2017arXiv170602262Z}. To the best of our knowledge, these works have focused mostly on generative scenarios  \cite{2017arXiv170602262Z}, and not on generic BNNs. The target of this paper is to leverage on these proposals to improve the training procedure of a Bayesian CNN and the estimation of the classification's uncertainty.

\subsubsection{Uncertainty quantification in BNNs}\label{subsec:uncertainty_quantification_in_bnns}
Quantifying the uncertainty of a prediction is a fundamental task in modern deep learning. In the context of BNNs, entropy allows to obtain a simple measure of uncertainty \cite{leibig2017leveraging}. The work in \cite{kendall2017uncertainties}, however, analyzed the difference between aleatoric uncertainty (due to the noise in the data), and epistemic uncertainty (due to volatility in the model specification) \cite{der2009aleatory,leibig2017leveraging,hullermeier2019aleatoric}. In order to properly model the former (which is not captured by standard entropy), they propose a modification of the BNN to also output an additional term necessary to quantify the aleatoric component. A further extension that does not require additional outputs is proposed in \cite{kwon2020uncertainty}. Their formulation, however, does not allow for a simple scalar definition in the multi-class case.

\section{Bayesian Neural Networks}

The core idea of Bayesian approaches is to estimate uncertainty using an entire distribution over the parameters, as opposed to the frequentist approach in which we estimate the solution of a problem as a fixed point. This is accomplished by using the Bayes' theorem: 
\begin{equation}
 \underbrace{p(\mathbf{w} \vert \mathcal{D} )}_{\text{posterior}} = \frac{p(\mathcal{D} \vert \mathbf{w}) p(\mathbf{w})}{p(\mathcal{D})}
 \propto \overbrace{p(\mathcal{D} \vert \mathbf{w} ) }^{\text{likelihood}}\underbrace{p(\mathbf{w})}_{\text{prior}}
 \label{eq:posterior}
\end{equation}
\noindent where $\mathcal{D}$ is a dataset and $\mathbf{w}$ is a set of parameters that we want to estimate. A BNN is a neural network with a distribution over the parameters specified according to \eqref{eq:posterior} \cite{neal2012bayesian}. In particular, we can see a DNN as a function $f_\mathbf{w}(x) = \bar{y}$ that, given a sample $x$ and parameters $\mathbf{w}$, computes the associated output $\bar{y}$. Bayesian methods give us the possibility to have a distribution of functions $p(\mathbf{w} \vert \mathcal{D})$ (the posterior) for a particular dataset $\mathcal{D}$, starting from a prior belief on the shape of the functions (the prior) and its likelihood on a single point, defined as $p(y \vert \mathbf{w}, x)$. 

Once we have the posterior, the inference step consists in integrating over all the possible configurations of parameters:
\begin{equation}
    p(y \vert x, \mathcal{D}) = \int p(y\vert \mathbf{w}, x) p(\mathbf{w} \vert \mathcal{D}) \ d\mathbf{w} \,.
    \label{eq:integrate_dw}
\end{equation}
%


\noindent This new equation represents a Bayesian Model Average (BMA): instead of choosing only one hypothesis (a single setting of the parameters $\mathbf{w}$) we, ideally, want to use any possible set of the parameters, weighted by the posterior probabilities. This process is called marginalization over the parameters $\mathbf{w}$. 









\subsection{Bayes by back-propagation}
\label{subsec:bbb}

In general the posterior in \eqref{eq:posterior} is intractable. As outlined in Section \ref{sec:related_work}, several techniques can be used to handle this intractability, and in this paper we focus on VI approximations, as described next. VI is an alternative to Markov Chain Monte Carlo (MCMC) methods, that can be used to faster approximate the posterior of Bayesian models if compared to MCMC but with less guarantees. For a complete review we refer to \cite{blei2017variational}. Generally speaking, the nature of BNNs (e.g., highly non-convex minimization problem, millions of parameters, etc.) makes these models very challenging for standard Bayesian methods.

Bayes By Back-propagation (BBB,  \cite{graves2011practical, blundell2015weight}) is a VI method to fit a variational distribution $q_\theta(\mathbf{w})$ with variational parameters $\theta$ over the true posterior, from which the weights can be sampled. The set $\theta$ of variational parameters can easily be found by exploiting the back-propagation algorithm, as shown afterward. This posterior distribution answer queries about unseen data points - given a sample and variational parameters $\theta$, taking the expectation with respect to the variational distribution. 
To make the process computationally viable, the expectation is generally approximated by sampling the weights from the posterior $T$ times; each set of weights gives us a DNN from which we predict the output, then the expectation is calculated as the average of all the $T$ predictions. Thus, Eq. \eqref{eq:integrate_dw} can be approximated as:
%
\begin{equation}
\label{eq:likelihood}
\begin{alignedat}{1}
    p(y \vert x, \mathcal{D}) & = \int p(y \vert \mathbf{w}, x) q_\theta(\mathbf{w}) d\mathbf{w} \\
    & \approx \frac{1}{T} \sum_{t=1}^T f_{\mathbf{w}_t}(x) \,.
\end{alignedat}
\end{equation}
%
\noindent where $\mathbf{w}_1, \ldots, \mathbf{w}_T$ are the $T$ sets of sampled weights. In the most common case, the variational family $q$ is chosen as a diagonal Gaussian distribution over the weights of the network. In this case, the variational parameters $\theta$ are composed, for each weight of the DNN, of a mean $\mu_i$ and a value $\rho_i$, which is used to calculate the variance of the parameter $\sigma_i = \log(1 +e^{\rho_i})$ (to ensure that the variance is always positive). Sampling a weight $w_i$ from the posterior is achieved by: $w_i = \mu_i + \log(1 + e^{\rho_i}) \odot \epsilon$, where $\epsilon \sim \mathcal{N}(0, 1)$; this technique is called re-parametrization trick \cite{blei2017variational}. Note that there exists alternative ways to codify the posterior over the parameters, and we explore some simplifications in the experimental section.

With this formulation, the parameters $\theta$ of the approximated posterior $q_\theta(\mathbf{w})$ can be found using the Kullback-Leibler (KL, \cite{kullback1951information}) divergence: 
\begin{align*}
    \theta^* & = \arg \min_\theta \text{KL} \big[ {q_\theta(\mathbf{w})} \Vert {p(\mathbf{w} \vert \mathcal{D})} \big]\\ 
    & = \arg \min_\theta \int q_\theta(\mathbf{w}) \log \frac{q_\theta(\mathbf{w})}{p(\mathbf{w}) p(\mathcal{D} \vert \mathbf{w})} \\
    & =  \arg \min_\theta \text{KL} \big[ {q_\theta(\mathbf{w})} \Vert {p(\mathbf{w})} \big]  - \mathbb{E}_{q_\theta(\mathbf{w})} \big[ \log p(\mathcal{D} \vert \mathbf{w} ) \big] \,.
\end{align*}
\noindent The optimal parameters $\theta^*$ are the ones that satisfy both the complexity of the dataset $\mathcal{D}$ and the prior distribution $p(\mathbf{w})$. The final objective function to minimize is: 
\begin{equation}
\begin{alignedat}{2}
        \mathcal{L}_{\text{ELBO}}(\theta)  & = &  \lambda( \mathbb{E}[q_\theta(\mathbf{w})] - \mathbb{E}[\log p(\mathbf{w})]) \\ 
        & & - \mathbb{E}_{q_\theta(\mathbf{w})} \big[ \log p(\mathcal{D} \vert \mathbf{w} ) \big]
\end{alignedat}
\label{eq:elbo}
\end{equation}
\noindent where $\lambda$ is an additional scale factor to weight the two terms. This equation is called ELBO because maximizing it is equivalent to minimizing the Kullback-Leibler divergence between the approximated posterior and the real one.
We can also look at this equation like as loss over the dataset plus a regularization term (the KL divergence). 

The ELBO function has limitations, one of them is that it might fail to learn an amortized posterior which correctly approximates the true posterior. This can happen in two cases: when the ELBO is minimized despite the fact that the posterior is inaccurate and when the model capacity is not sufficient to achieve both a good posterior as well as good data fitting. For further information we refer to \cite{alemi2017fixing} and \cite{2017arXiv170602262Z}. 





\subsection{Measuring the uncertainty of a prediction}
\label{subsec:measuring_uncertainty}

As introduced in Section \ref{subsec:uncertainty_quantification_in_bnns}, the uncertainty of a prediction vector $\mathbf{p}$ can be calculated in many ways. In a classification setting, where vector $\mathbf{p}$ contains the probability associated to the classes, the most straightforward way is the entropy: 
\begin{equation}
    H(\mathbf{p}) = - \mathbf{p}^T \log(\mathbf{p}) \,,
\end{equation}
\noindent 
Combining this formulation with Eq. \eqref{eq:likelihood}, the classification entropy can be calculated as:
\begin{equation}
    \label{eq:entropy}
    H(p(y \vert \mathcal{D}, x)) =\frac{1}{T} \sum_{t=1}^T - \mathbf{\hat{p}}_t^T \log \mathbf{\hat{p}}_t
\end{equation}
\noindent where $\mathbf{\hat{p}}_t = \text{softmax}(f_{\mathbf{w}_t}(x))$ and $T$ is the number of weights sampled from the posterior. 
This entropy formulation allows the calculation of the uncertainty also for a BNN, 
However, a more suitable measure of uncertainty, exploiting the possibility of sampling the weights $\mathbf{w}$ to calculate the cross uncertainty between the classes (a covariance matrix), can be formulated \cite{kwon2020uncertainty}. To this end, we define the variance of the predictive distribution \eqref{eq:likelihood} as:

%
\begin{equation}
\label{eq:var}
\begin{split}
    \text{Var}[y] &= \mathbb{E}[y y^T] - \mathbb{E}[y]\mathbb{E}[y]^T  \\ 
    & = \int \big[ \text{diag}(\mathbb{E}[y] ) - \mathbb{E}[y]\mathbb{E}[y]^T \big] q_\theta(\mathbf{w}) d \mathbf{w}\\
    & + \int \big[ (\mathbb{E}[y]- \mathbb{E}[\hat{y}]) (\mathbb{E}[y]- \mathbb{E}[\hat{y}])^T \big] q_\theta(\mathbf{w}) d \mathbf{w}
\end{split}
\end{equation}
\noindent where $\hat{y} = p(y \vert x, \mathcal{D})$ and $y = p(y \vert \mathbf{w}, x)$. For further information about the derivation, we refer to \cite{kwon2020uncertainty}. The first term in the variance formula is called aleatoric uncertainty, while the second one is the epistemic uncertainty \cite{der2009aleatory}. The first quantity measures the inherent uncertainty of the dataset $\mathcal{D}$, it is not dependent on the model, and more data might not reduce it, instead the second term incorporates the uncertainty of the model itself, and can be decreased by augmenting the dataset or by redefining the model. 
In \cite{kendall2017uncertainties} and \cite{kwon2020uncertainty} the authors have proposed different ways to approximate these quantities. In \cite{kendall2017uncertainties} the authors constructed a BNN and
used the mean $\alpha$ and the standard deviation $\beta$ of the logits, the output of the last layer before the softmax activation function, to calculate the variance:
\begin{align*}
    \text{Var}[p(y \vert \mathcal{D}, x)] &= \frac{1}{{T}} \sum_{t=1}^{T} \text{diag}(\beta_t^2) & \text{(aleatoric)}\\
    & + \frac{1}{{T}}  \sum_{t=1}^{T} ( \alpha_t - \bar{\alpha}_t) ( \alpha_t - \bar{\alpha}_t)^T & \text{(epistemic)}
\end{align*}
\noindent where $\mathbf{\bar{\alpha}} = \sum_t \alpha_t/T$. In \cite{kwon2020uncertainty} the authors highlighted the problems of this approach: it models the variability of the logits (and not the predictive probabilities), ignoring that the covariance matrix is a function of the mean vector; moreover, the aleatoric uncertainty does not reflect the correlation due to the diagonal matrix modeling. To overcome these limitations, they proposed an improvement: 
\begin{equation} 
\label{eq:var1}
\begin{alignedat}{2}
    \text{Var}[p(y \vert \mathcal{D}, x)] &= \frac{1}{T} \sum_{t=1}^T \text{diag}(\mathbf{\hat{p}}_t) - \mathbf{\hat{p}}_t \mathbf{\hat{p}}_t^T & \quad & \ \text{(aleatoric)}  \\ 
    & + \frac{1}{T}  \sum_{t=1}^T ( \mathbf{\hat{p}}_t - \mathbf{\bar{p}}_t) ( \mathbf{\hat{p}}_t - \mathbf{\bar{p}}_t)^T & \quad & \ \text{(epistemic)} 
\end{alignedat}
\end{equation}
\noindent where $ \mathbf{\bar{p}}_t = \sum \mathbf{\hat{p}}/T$. 
This formulation converges in probability to Eq. \eqref{eq:var} as the number of samples $T$ increases. In the case of binary classification, the formula simplifies to:
\begin{equation} 
\label{eq:var2}
\begin{alignedat}{2}
    \text{Var}[p(y \vert \mathcal{D}, x)] = &\mathbb{E}[\mathbf{\hat{p}}_t^2] - \mathbb{E}[\mathbf{\hat{p}}_t]^2 \ \ &\text{(aleatoric)} +\\
    & \mathbb{E}[\mathbf{\hat{p}}_t(1-\mathbf{\bar{p}}_t)] \ \ &\text{(epistemic)}
\end{alignedat}
\end{equation}
\noindent This definition is more viable because it calculates a scalar instead of a matrix, but cannot be used trivially if the problem involves more than two classes; if not by collapsing all the probabilities that are less than the maximum one into one single probability and treating the problem as a binary one. In this paper, we also present a modified version of the definition \eqref{eq:var1}, which can be used to evaluate the uncertainty as a scalar also in multiclass scenarios.

\section{Proposed approaches} 

In this section, we introduce the proposed variations for the training of BNNs. Firstly, we outline a new way to approximate the weights' posteriors, leading to a better posterior approximation, higher accuracy, and an easier minimization problem. Secondly, we provide an improvement of the measure of uncertainty \eqref{eq:var1}, which is more suited for problems that are not binary classification tasks.

\subsection{Posterior approximation via Maximum Mean Discrepancy regularization}

\begin{algorithm}[t]
\caption{One epoch of the training procedure.}
\label{algo:proposed_approach}
\begin{algorithmic}[1]
\State Given a prior distribution $p(\mathbf{w})$, a kernel function $\kappa$, the set of variational parameters $\bm{\theta} = (\bm{\mu}, \bm{\rho})$, a loss function $\mathcal{L}_d$, and the number of sampling steps $T$.

\For{each batch $(x, y)$ in the dataset}
\State $\mathcal{L} \gets 0$

\For{$i =0 \dots T$}
\State $\mathbf{w^q} = \bm{\mu} + \log(1 + e^{\bm{\rho}}) \odot \bm{\epsilon}$ 
\Comment{with $\bm{\epsilon} \sim \mathcal{N}(0, 1)$}
\State $\mathbf{w}^p \sim p(\mathbf{w})$

\State $\mathcal{L}_{\text{MMD}} \gets \overline{\text{EMMD}}^2_\kappa \big(\mathbf{w}^{q}, \mathbf{w}^{p})$
\State $\mathcal{L} \gets \mathcal{L} +\lambda \mathcal{L}_{\text{MMD}}  + \mathcal{L}_d(f_{\mathbf{w}^q}(x), y) $
\EndFor
\State $\mathcal{L} \gets \mathcal{L} / T$
\State update $\bm{\theta}$ using $\Delta \frac{\partial \mathcal{L}}{\partial \mathbf{w}}$ 
\EndFor
\label{euclidendwhile}

\end{algorithmic}
\end{algorithm}

The MMD estimator was originally introduced as a non-parametric test for distinguishing samples from two separate distributions \cite{Gretton:2012:KTT:2188385.2188410}. It can be used to build an universal estimator \cite{cherief2019finite} which is robust to outliers, and also in the Bayesian statistics \cite{abdellatif20a} and to build a generative model \cite{briol2019statistical}. 
Formally, denote by $z$ and $z'$ two samples from an independent random variable with distribution $\mathrm{Z}$, by $v$ and $v'$ two samples from an independent random variable with distribution $\mathrm{V}$, and by $\kappa$ a characteristic positive-definite kernel. The square of the MMD distance between the two distributions is defined as:
\begin{align*}
    \text{MMD}^2_\kappa(\mathrm{Z}, \mathrm{V}) &= \lVert\mu_\mathrm{Z} - \mu_\mathrm{V}\rVert
    = \\ 
&=  \mathbb{E}[\kappa(z, z')] + 
\mathbb{E}[\kappa(v, v')] 
-2 \mathbb{E}[\kappa(z, v)] \,.
\end{align*}
\noindent We have that $\text{MMD}_\kappa(\mathrm{Z}, \mathrm{V}) = 0 \iff \mathrm{Z} = \mathrm{V}$. Inspired by \cite{2017arXiv170602262Z} and \cite{li2017mmd} (who considered a similar approach for generative models), we propose to replace the KL term in \eqref{eq:elbo} with an MMD estimator, i.e., we propose to search for a variational set of parameters $\theta$ that minimizes the MMD distance with respect to the prior $p(\mathbf{w})$:
\begin{align}
\theta^* =  \arg \min_\theta  \Big\{ \text{MMD}^2_{\kappa} & \big( {q_\theta(\mathbf{w})}, {p(\mathbf{w})} \big)  - \nonumber\\ & \mathbb{E}_{q_\theta(\mathbf{w})} \big[ \log p(\mathcal{D} \vert \mathbf{w} ) \big] \Big\} \,.
\label{eq:proposed_approach}
\end{align}
In practice, the quantity $\text{MMD}^2_{\kappa} \big( {q_\theta(\mathbf{w})}, {p(\mathbf{w})} \big)$ can be estimated using finite samples from the two distributions.
Given a sample $\mathbf{w}^q \sim q_\theta(\mathbf{w})$ and a sample $\mathbf{w}^p \sim p(\mathbf{w})$, an unbiased estimator of $\text{MMD}^2_{\kappa} \big( {q_\theta(\mathbf{w})}, {p(\mathbf{w})} \big)$ is given by:
\begin{equation}
\label{eq:mmd}
\begin{alignedat}{1}
     \text{EMMD}^2_\kappa(\mathbf{w}^q, \mathbf{w}^p) &= \frac{1}{n(n-1)}\sum_{i \ne j} 
    \kappa({w}^q_i, {w}^q_j) \\ 
    &+ \frac{1}{n(n-1)}\sum_{i \ne j} \kappa({w}^p_i, {w}^p_i) \\
    &- \frac{2}{n^2}\sum_{i, j} \kappa({w}^q_i, {w}^p_j)
\end{alignedat}
\end{equation}
\noindent where $w_i^p$ is the $i$th element of $\mathbf{w}^p$ (and similarly for $\mathbf{w}^q$), and both vectors have size $n$. Using the unbiased version, the results can be negative if the two distributions are very close to each other. For this reason we use a different formulation in which, to speed up the convergence, we decide to eliminate the negative part: 
\begin{equation*}
    \overline{\text{EMMD}}^2_\kappa \big( \mathbf{w}^q, \mathbf{w}^p \big) = \max \big(0,  \text{EMMD}^2_\kappa(\mathbf{w}^q, \mathbf{w}^p) \big)
\end{equation*}
%

\noindent As stated earlier, the idea of using the MMD distance connected to neural network models was originally explored in \cite{2015arXiv150503906K} and \cite{DBLP:journals/corr/LiSZ15}, who were mostly focused on generative models.  
The power of minimizing the MMD distance relies on the fact that it is equivalent to minimizing a distance between all the moments of the two distributions, under an affine kernel. In \eqref{eq:proposed_approach} we use this metric as a regularization approach, which minimizes the distance between the posterior over the parameters $\theta$ and the chosen prior. Summarizing, we propose to estimate the posterior by minimizing: 
\begin{equation}
\begin{alignedat}{1}
\mathcal{L}_{\text{MMD}}(\theta) & = \frac{1}{T} \sum_{t=1}^T \Big[\lambda \overline{\text{EMMD}}^2_\kappa \big(\mathbf{w}^{q_t}, \mathbf{w}^{p_t})
  - \\ &  \log p(\mathcal{D} \vert \mathbf{w}^{q_t} ) \Big]
\end{alignedat}
\end{equation}
\noindent with $\mathbf{w}^{q_i} \sim q_\theta(\mathbf{w})$ and $\mathbf{w}^{p_i} \sim p(\mathbf{w})$, $T$ the number of times that we sample from the two distributions, while the value $\lambda$ is an additional scale factor to balance the classification loss and the posterior's one; as in \cite{blundell2015weight}, we set $\lambda$ to $\lambda(i; B)= \frac{2^{B-i}}{2^B - 1}$, where $i$ is the current batch in the training phase and $B$ is the total number of batches; in this way, the first optimization steps are influenced by the prior more than the future ones, which are influenced only by the data samples. The pseudocode for training our model is summarized in Algorithm \ref{algo:proposed_approach}.

    
\subsection{Bayesian Cross Uncertainty (BCU)} 
In this section, we propose a modified version of the uncertainty measure formulated in Eq. \eqref{eq:var1}, that we call Bayesian Cross Uncertainty (BCU).

The variance formulated in \eqref{eq:var1} gives us a $c \times c$ matrix, with $c$ the number of classes of our classification problem. Sometimes, it is useful to have a scalar value, which indicates the uncertainty of our prediction and that can be easily used or visualized, so that different uncertainties (or models) can be compared easily. 

The most straightforward approach to reduce a matrix to a scalar is to calculate its determinant, the sparser the matrix is the closer the resulting value will converge to zero. This approach comes with an inconvenience: in a binary classification problem, if we have, for a sample $x$, two vectors of predictions $\mathbf{p}_1 = [1, 0]$, which codify the absolute certainty of the prediction, and $\mathbf{p}_2 = [0.5 , 0.5]$, indicating that the network is maximally uncertain, we have that $\big\vert\text{Var}[\mathbf{p}_1]\big\vert = \big\vert\text{Var}[\mathbf{p}_2]\big \vert = 0$. To avoid these cases, we propose to modify the formulation of Eq. \eqref{eq:var1} as follows: 
\begin{equation}
    \overline{\text{Var}}[p(y \vert \mathcal{D}, x)] = \text{Var}[p(y \vert \mathcal{D}, x)] + \frac{1}{c}\mathbf{I}_c
    \label{eq:unnormalized_bcu}
\end{equation}
\noindent where $c$ is the number of classes and $\mathbf{I}_c \in \mathbb{R}^{c \times c}$ is the identity matrix. In this case, we have that the determinant of Eq. \eqref{eq:var_hat} is lower bounded when we have utmost confidence, and this bound is equal to the determinant of the matrix $\frac{1}{c}\mathbf{I}_c$: $\inf(\big\vert\overline{ \text{Var}}[p(y \vert \mathcal{D}, x)]\big\vert)=\vert\frac{1}{c}\mathbf{I}_c\vert=c^{-c}$. To calculate the upper bound we need to study when such a scenario could emerge. The possible scenarios in which we have the utmost uncertainty are the following: in the first one the network produces the same probability, $\frac{1}{c}$, for each class (utmost aleatoric uncertainty), while in the second one we have a sample that is classified $T$ times, with $T=c$, and at each prediction the network assign a probability equals to $1$ to a different class, and zeros to the others (utmost epistemic uncertainty). 
In these cases, the upper bound is: $\sup\big(\big\vert\overline{\text{Var}}[p(y \vert \mathcal{D}, x)]\big\vert\big)=2^{c-1} c^{-c}$. These two values can be used to normalize the result of Eq. \eqref{eq:var_hat} between zero, maximum certainty, and one, utmost uncertainty, given that this formulation ensures a bounded measure of uncertainty. In this way, the uncertainty is well defined for a BNN model, since it reaches its maximum only when one of two terms, epistemic or aleatoric, reaches it. The final measure of uncertainty that we propose is the normalized version of \eqref{eq:unnormalized_bcu}:
\begin{equation}
\label{eq:var_hat}
    \text{BCU}[p(y \vert \mathcal{D}, x)] = \frac{\Big\vert \overline{\text{Var}}[p(y \vert \mathcal{D}, x)] \Big\vert - u_{\text{min}}}{(u_{\text{max}} - u_{\text{min}})} \,,
\end{equation}
where $u_{\text{min}}$ and $u_{\text{max}}$ are the minimum and maximum values as defined above. Furthermore, we define a way of discarding a sample based on its classification's uncertainty. When the training of the DNN is over, we collect all the measures of uncertainty associated to the samples that have been classified correctly in a set that we call $\chi$. From this set of uncertainties $\chi$, we define a threshold as:
\begin{equation}
\label{eq:thres}
    \text{Threshold}_\gamma(\chi) = \text{Q3}(\chi) + \gamma \big[\text{Q3}(\chi) - \text{Q1}(\chi)\big]
\end{equation}
\noindent where $\text{Q1}(\chi)$ and $\text{Q3}(\chi)$ are, respectively, two functions that return the first and the third quartile of the set $\chi$, and $\gamma$ is an hyper-parameter. Once a threshold is calculated, a new sample can be discarded if its associated uncertainty exceeds it. 

We underscore that this way of discarding images is not related to the formulation of variance in Eq. \eqref{eq:var} or the BCU, nor to the BNNs, but can be used with every combination of DNN and measures of uncertainty.

\section{Neural networks calibration}

BNNs are more suitable for a real world decision making application, due to the possibility to give an interval of confidence for the prediction, as explored in Section \ref{subsec:measuring_uncertainty}. However, another important aspect in these scenarios, apart from the correctness of the predictions, is the ability of a model to provide a good \textit{calibration}: 
the more the network is confident about a prediction, the more the probability associated with the predicted class label should reflect the likelihood of a correct classification.

In \cite{niculescu2005predicting} the authors proved that shallow neural networks are typically well calibrated for a binary classification task. On the other hand, when considering deeper models, while the networks' predictions become more accurate, due to the growing complexity, they also become less 
calibrated, as pointed out in \cite{guo2017calibration}. In this work, we also analyze how calibrated BNNs are. 
In particular, we show in the experimental section that the proposed MMD estimator leads to better calibrated models.

Given a sample $x$, the associated ground truth label $y$, the predicted class $\hat{y}$ with its associated probability of correctness $\hat{p}$, we want that: 
\begin{equation}
\label{eq:calibration}
    p(\hat{y} = y \ \vert \ \hat{p} = p) = p \ \ \forall \ p \in [0, 1] \,.
\end{equation}
\noindent This quantity cannot be computed with a finite set of samples, since $\hat{p}$ is a continuous random variable, but it can be approximated and visually represented (as proposed in \cite{degroot1983comparison} and \cite{niculescu2005predicting}) using the following formula: 
\begin{equation}
\label{eq:acc}
    \text{acc}(B_m) = \frac{1}{\vert B_m \vert} \sum_{i \in B_m} \mathbb{1}(\hat{y}_i = y_i)
\end{equation}
\noindent where $\hat{y}_i$ and $y_i$ are the predicted and the true label for the sample $i$, and, chosen the number $M$ of splits of the range $[0, 1]$ (each one has size equals to $M^{-1}$), we group the predictions into $M$ interval bins $B_m$. Each $B_m$ is the set of indices of samples with a prediction confidence that falls into the range $( \frac{m-1}{M}, \frac{m}{M}]$. The Eq. \eqref{eq:acc} can be combined with a measure of confidence calculated as:
\begin{equation*}
\label{eq:conf}
    \text{conf}(B_m) = \frac{1}{\vert B_m \vert} \sum_{i \in B_m} \hat{p}_i
\end{equation*}
\noindent to understand if a model is calibrated, which is true when $\text{acc}(B_m) = \text{conf}(B_m)$, for each bin $B_m$ with $m \in \{1, \dots, M\}$. Not only these formulas provide a good visualization tool, namely reliability diagram, but also it is useful to have a scalar value which summarizes the calibration statistics. The metric that we use is called Expected Calibration Error (ECE, \cite{naeini2015obtaining}): 
\begin{equation}
\label{eq:ECE}
    \text{ECE} = \sum_{m=1}^M \frac{\vert B_m \vert}{n}  \big| \text{acc}(B_m) - \text{conf}(B_m)\big|
\end{equation}
\noindent where $n$ is the total number of samples. The resulting scalar gives us the calibration gap between a perfectly calibrated network and the evaluated one.

\section{Experiments}

To evaluate the proposed VI method, we start with a toy regression task for visualization purposes, before moving to different datasets for image classification. We compare our proposed model with others state-of-the-art approaches. 
We put particular emphasis on evaluating different priors and seeing how this choice affects the final results (robustness), to study the calibration of BNNs, and how well our measure of uncertainty behaves when we want to discard images on which the network is uncertain (e.g., adversarial attacks). 
The code to replicate the experiments can be found in a public repository.\footnote{\url{https://github.com/ispamm/MMD-Bayesian-Neural-Network}}

\subsection{Case study 1: Regression}
\label{subsec:case_study_1_regression}
\begin{figure*}[t!]
\centering
  \begin{subfigure}[b]{0.23\textwidth}
    \includegraphics[width=\textwidth]{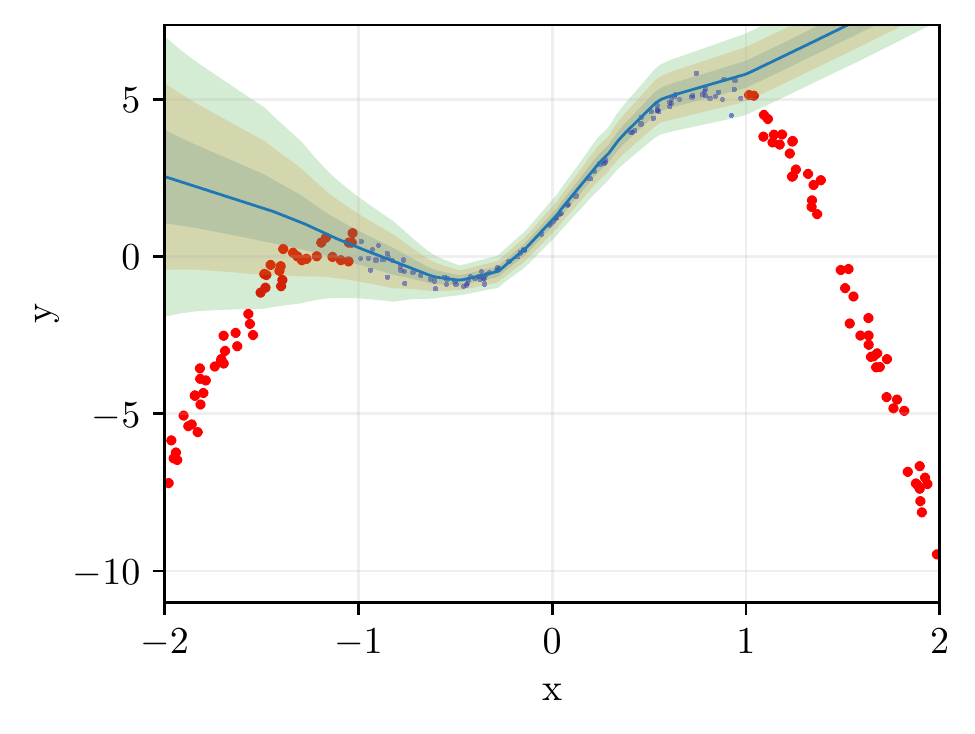}
    \caption{DNN}
  \end{subfigure}
  \begin{subfigure}[b]{0.23\textwidth}
    \includegraphics[width=\textwidth]{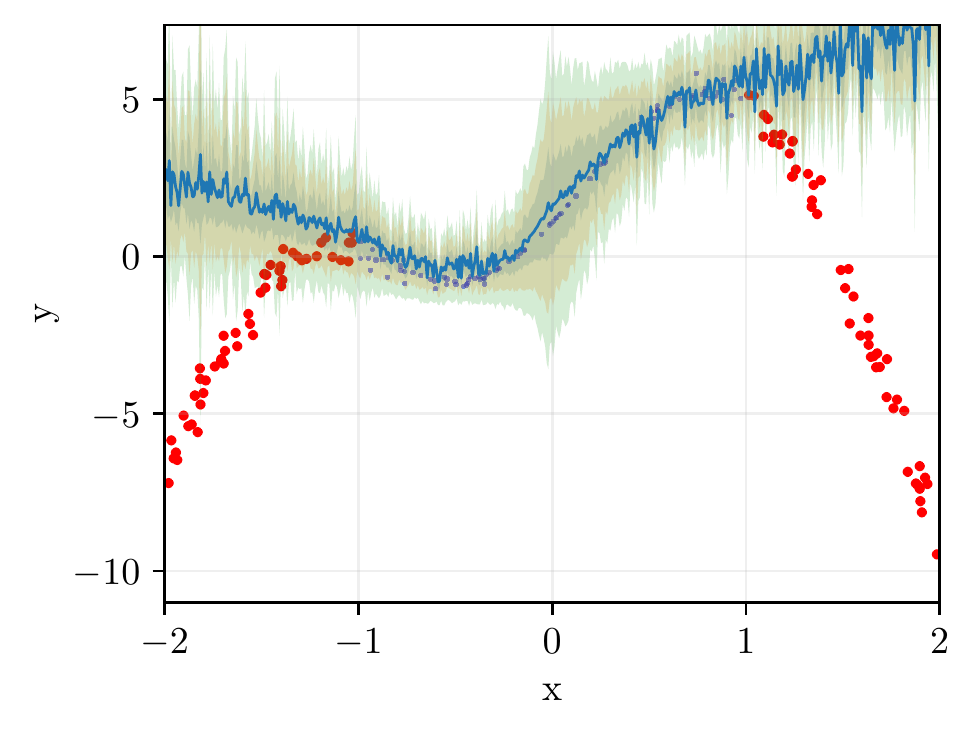}
    \caption{MC Dropout}
  \end{subfigure}
\begin{subfigure}[b]{0.23\textwidth}
    \includegraphics[width=\textwidth]{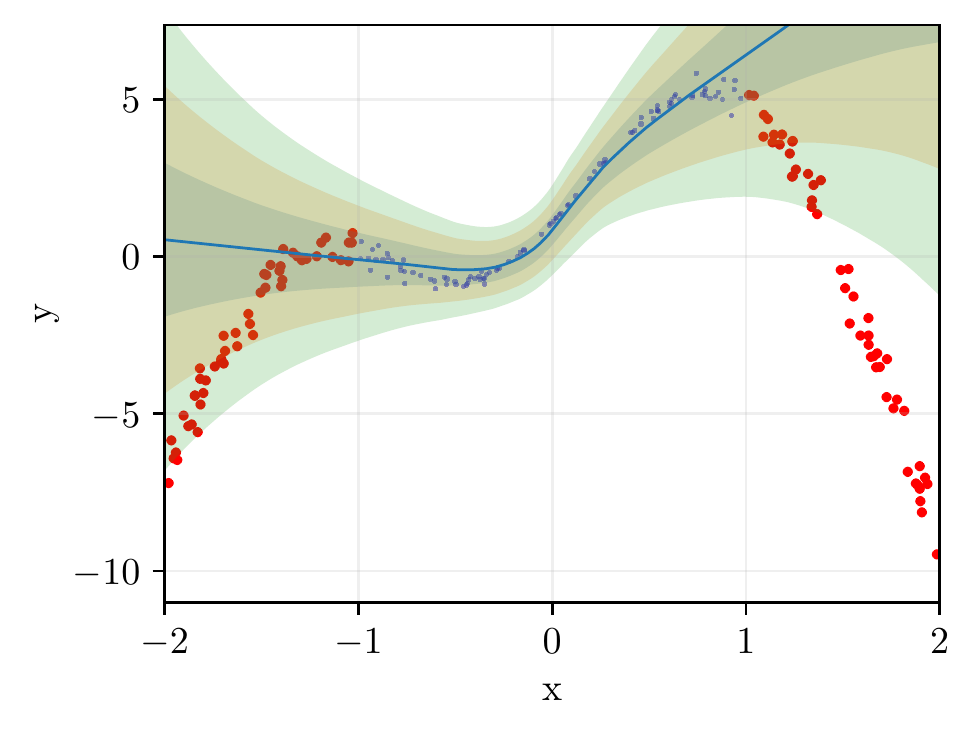}
    \caption{BBB}
  \end{subfigure}
  \begin{subfigure}[b]{0.23\textwidth}
    \includegraphics[width=\textwidth]{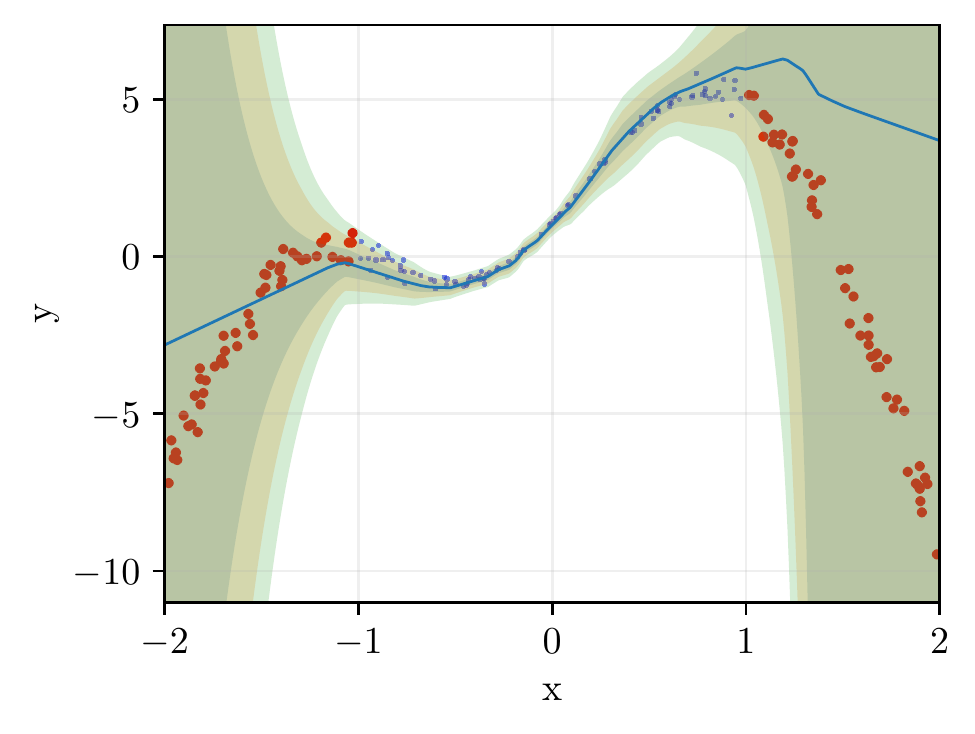}
    \caption{MMD}
  \end{subfigure}
  \caption{The images show the results obtained on the heteroscedastic regression problem with additive noise equals to $0.3$. The line represents the prediction, the smaller points are the train dataset, while the bigger ones are test points outside the training range, to check how the function evolves; we also show the variances of the prediction.  
  }
  \label{img:hetero_reg}
\end{figure*}
\begin{figure*}[h!]
\centering
  \begin{subfigure}[b]{0.23\textwidth}
    \includegraphics[width=\textwidth]{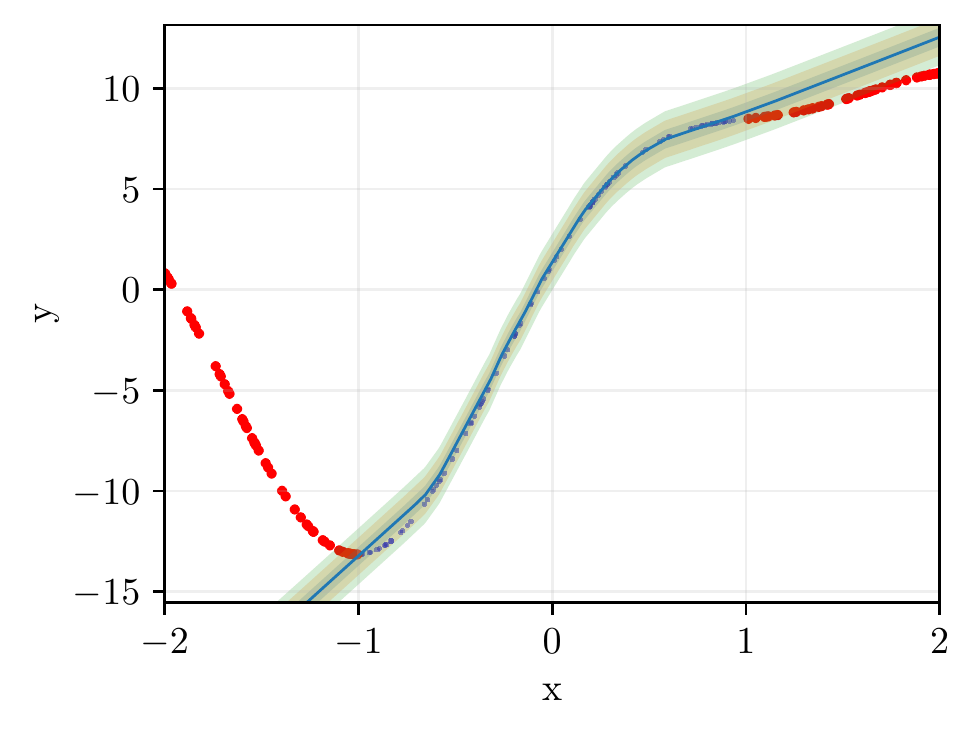}
    \caption{DNN}
  \end{subfigure}
  \begin{subfigure}[b]{0.23\textwidth}
    \includegraphics[width=\textwidth]{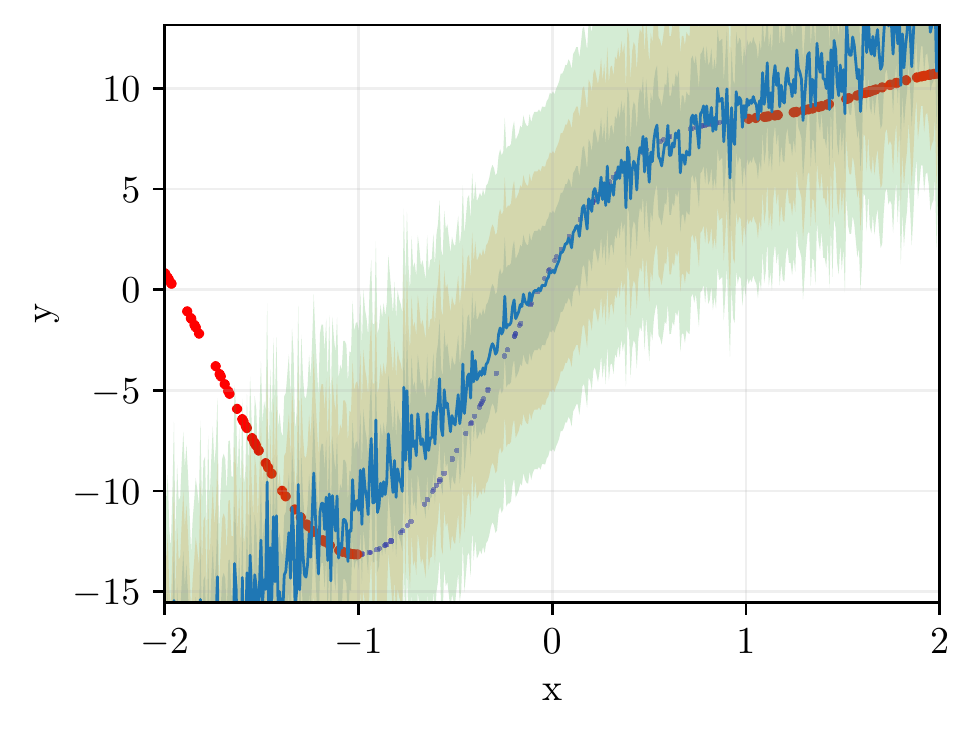}
    \caption{MC Dropout}
  \end{subfigure}
\begin{subfigure}[b]{0.23\textwidth}
    \includegraphics[width=\textwidth]{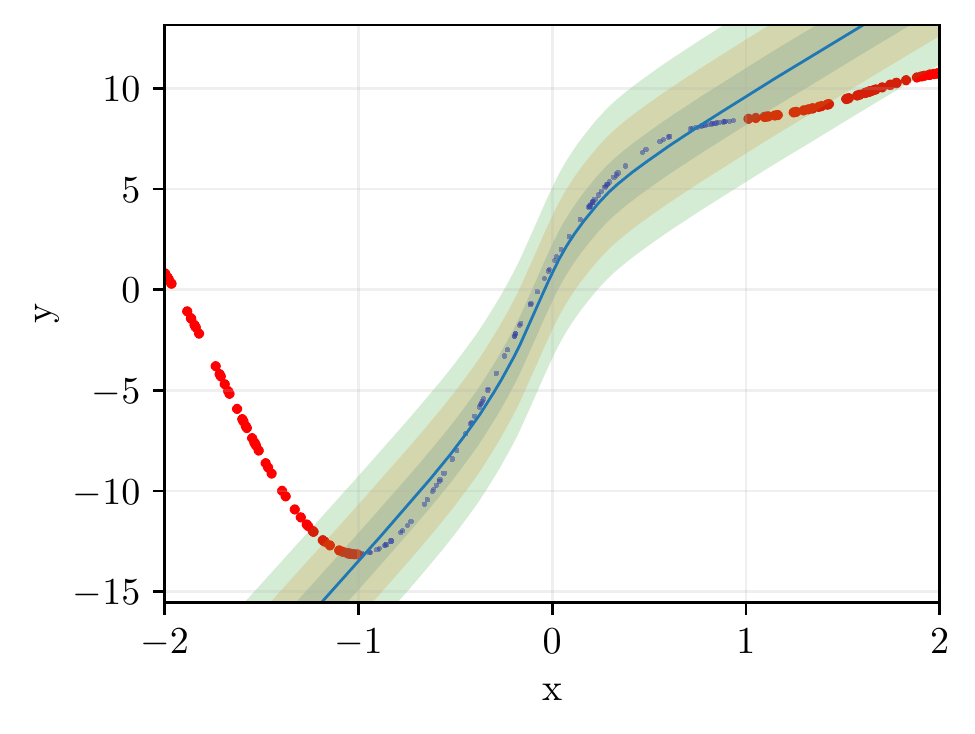}
    \caption{BBB}
  \end{subfigure}
  \begin{subfigure}[b]{0.23\textwidth}
    \includegraphics[width=\textwidth]{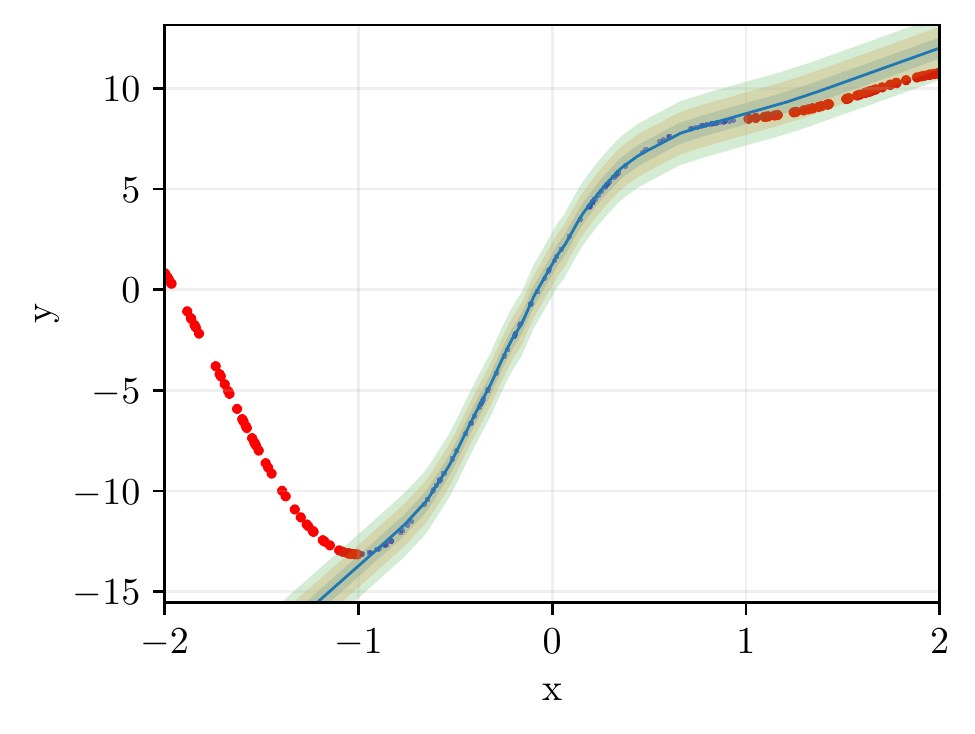}
    \caption{MMD}
  \end{subfigure}
  \caption{The images show the results obtained on the hemoroscedastic regression problem with additive noise equals to $0$. The line represents the prediction, the smaller points are the train dataset, while the bigger ones are test points outside the training range, to check how the function evolves; we also show the variances of the prediction. }
  \label{img:homo_reg}
\end{figure*}
In this section, we evaluate the models on a toy regression problem, in which the networks should learn the underlying distribution of the points, and then being able to provide reasonable predictions even in regions outside the training one. 

Each regression dataset is generated randomly using a Gaussian Process, with the RBF kernel, given a range in which the points lie, the number of points to generate and the variance of the additive noise. We generated two different kinds of regression problems: homoscedastic and heteroscedastic. In the first one, the variance is shared across all the random variables, while in the second one each random variable has its own. 
Each experiment consists in 100 training points in the range $[-1, 1]$ and 100 testing points outside this range. In the MMD experiments, we used the RBF kernel $\kappa(x; A) = \sum_{\gamma \in A} \text{exp}\big\{\frac{\lVert x\rVert^2}{2\gamma^2}\big\}$  with $A = \{0.5, 1, 2, 4, 8, 16\}$ to regularize the posterior. 


We compare our approach with a standard DNN, the BBB method and a network which uses the dropout layer to approximate the variational inference (proposed in \cite{gal2015dropout}) by keeping the dropout turned on even during the test phase. This technique is called Monte Carlo Dropout (MC Dropout). For all the experiments we trained the network for 100 epochs using RMSprop with a learning rate equal to $0.01$.

As prior for BBB and MMD we used a Gaussian distribution $p(\mathbf{w}) = \mathcal{N}(0, 10)$. We initialize the $\mu$ as proposed in \cite{he2015delving} and $\rho \sim \mathcal{U}(-10, -6)$, to keep the resulting weights around a value that guarantees the convergence of the optimization procedure. In these experiments we do not vary the prior distribution.

In Fig. \ref{img:hetero_reg} we show the results obtained from the heterocedastic experiment with additive noise equals to $0.3$. We can see that the BNN trained with MMD is the only one capable of reasonably estimating the interval of confidence in regions outside the training one. While the DNN and MC Dropout are too confident about their predictions, BBB gives the less confident predictions, but we can see that it fails to understand that the uncertainty should increase outside the training range. In Fig. \ref{img:homo_reg} the results obtained on a homoscedastic regression problem are shown, with similar trends.

\subsection{Case study 2: Image classification}
\label{subsec:image_classification}
\begin{table*}[t!]
\centering
\caption{The table shows, for each method, the accuracy results and the associated standard deviation, both expressed in percentage, obtained on the classification benchmarks. Some results are missing because no combination of parameters lead to convergence of the classification task.}
\resizebox{\textwidth}{!}{%
\begin{tabular}{c|c|c|c|c|c|c|}
\cline{2-7}
\multicolumn{1}{l|}{}          & \multirow{2}{*}{DNN} & \multirow{2}{*}{MC Dropout} & \multicolumn{2}{c|}{BBB}             & \multicolumn{2}{c|}{MMD}                \\ \cline{4-7} 
\multicolumn{1}{l|}{} &
   &
   &
  \multicolumn{1}{l|}{Neuron-wise} &
  \multicolumn{1}{l|}{Weight-wise} &
  \multicolumn{1}{l|}{Neuron-wise} &
  \multicolumn{1}{l|}{Weight-wise} \\ \hline
\multicolumn{1}{|c|}{MNIST}    & 98.59 $\rpm$ 0.08    & 98.30 $\rpm$ 0.09            & 98.16   $\rpm$ 0.02         & 28.17 $\rpm$ 2.69 & 98.64 $\rpm$ 0.061 & \bt 98.84 $\rpm$ 0.02 \\ \hline
\multicolumn{1}{|c|}{CIFAR10}  & 74.73 $\rpm$ 0.36    & 75.56 $\rpm$ 0.01        & 65.73 $\rpm$ 0.50 & -                 & 75.24 $\rpm$ 0.29  & \bt 75.64 $\rpm$ 0.12  \\ \hline
\multicolumn{1}{|c|}{CIFAR100} & 39.89 $\rpm$ 0.33    & 38.85 $\rpm$ 0.20            & 35.31 $\rpm$ 038 & -                 & 42.2 $\rpm$ 0.51   & \bt 42.36 $\rpm$ 0.36   \\ \hline
\end{tabular}%
}
\label{table:score}
\end{table*}

In this section, we present the results obtained on image classification experiments. To the best of our knowledge, no competitive results on this field have been proposed using BNNs; the best results are present in \cite{2019arXiv190102731S}, in which the authors used the local re-parametrization trick, described in \cite{kingma2015variational}: a technique in which the output of a layer is sampled instead of the weights. The main problem of this technique applied to the CNNs is that it doubles the number of operations inside a layer (e.g., in the CNN case we have two convolutions, one for the mean and the other for the variance of the layer's output). For this reason, we believe that it is not computationally reasonable, especially with deeper architectures, and MMD could be a step towards better Bayesian CNNs. 

Our main concern is to show that the MMD approach works even with a ``bad" prior, which implies having small knowledge about the problem. For this purpose, we studied different priors: the Gaussian distribution $\mathcal{N}(0, \sigma)$, the Laplace distribution $\text{Laplace}(0, b)$, the uniform distribution $\mathcal{U}(-a, a)$ and the Scaled Gaussian Mixture $\mathcal{SN}(\sigma_1, \sigma_2, \pi)$ from \cite{blundell2015weight}. In addition, we will evaluate the introduced measure of uncertainty under the Fast Gradient Sign Method (FGSM, \cite{goodfellow2014explaining}). Finally, we evaluate the calibration of each network. 

To this end, we evaluate the methods on three datasets: the first is MNIST \cite{lecun2010mnist}, the second one is CIFAR10, and the last one is a harder version of CIFAR10 called CIFAR100, which contains the same number of images, but 100 classes instead of 10. 
For all the experiments, we used the Adam optimizer \cite{kingma2014adam} with the learning rate set to $1e-3$, and the weights initialized as in the regression experiments, to ensure a good gradient flow. For MNIST, we used a simple network composed by one CNN layer, with 64 kernels, followed by max pooling and two linear layers. For CIFAR10, we used a network composed by three blocks of convolutions and max pooling, respectively with $64$, $128$ and $256$ kernels, followed by three linear layers; for CIFAR100, we used the same architecture but doubling the number of kernels. In all the architectures, the activation function is the ReLU.

We trained all the networks for 20 epochs; we also implemented an early stopping criteria, in which training is stopped if the validation score does not improve for 5 consecutive epochs.
For BBB, MMD, and MC Droput we sampled one set of weights during the train phase and 10 sets during the test phase. 
To have better statistics of the results, we repeated each experiment $5$ times.

Since the posterior over the weights doubles the number of parameters, leading to a minimization problem which is harder to minimize, we decided also to test a simplification of it, called neuron-wise posterior. This posterior is defined as $\mathcal{N}_\theta(\mu_{i,j,k}, \alpha_{i,j}\mu^2_{i,j,k}
)$, in which each weight connecting neuron $i$ to neuron $j$ in layer $k$ has its own mean, but the variance is given by $\mu^2_{i,j,k}$ scaled by a parameter $\alpha_{i,j}$ which is defined neuron-wise.  In this way, we have less parameters and the minimization problem could benefit from it.

\subsubsection{Prior choice}
\label{subsec:prior_robustness}
\begin{table}[!t]
\centering
\caption{The Table shows the accuracy results, on CIFAR10, about the robustness w.r.t. the prior choice. Some results are missing because no combination of parameters lead to convergence of the classification task.}
\label{table:priors}
\resizebox{0.8\columnwidth}{!}{%
\begin{tabular}{|c|c|c|c|}
\hline
\multirow{2}{*}{Prior}           & \multicolumn{1}{c|}{BBB} & \multicolumn{2}{c|}{MMD} \\ \cline{2-4} 
                                 & Neuron-wise                  & Neuron-wise     & Weight-wise    \\ \hline \hline
$\mathcal{N}(0, 1)$              & 66.26                    & 75.43       & 75.43      \\ \hline
$\mathcal{N}(0, 0.1)$            & 33.28                    & 74.59       & 75.04      \\ \hline
$\mathcal{N}(0, 0.01)$           & -                        & 75.47       & 75.64      \\ \hline \hline
$\text{Laplace}(0, 1)$           & 52.84                    & 75.47       & 75.32      \\ \hline
$\text{Laplace}(0, 0.1)$         & 12.11                    & 74.90        & 75.30       \\ \hline
$\text{Laplace}(0, 0.01)$        & -                        & 75.58       & 75.47      \\ \hline
$\text{Laplace}(0, 0.05)$        & -                        & 74.46       & 74.89      \\ \hline \hline
$\mathcal{U}(-1, 1)$             & 66.23                    & 74.67       & 75.70       \\ \hline
$\mathcal{U}(-0.1, 0.1)$         & 66.23                    & 75.6        & 75.93      \\ \hline
$\mathcal{U}(-0.2, 0.2)$         & 66.23                    & 74.95       & 74.89      \\ \hline \hline
$\mathcal{SN}(1, 0.1, 0.5)$      & 66.23                    & 74.67       & 75.70       \\ \hline
$\mathcal{SN}(1, 0.1, 0.2)$      & 66.23                    & 75.60       & 75.93      \\ \hline
$\mathcal{SN}(0.1, 0.001, 0.5)$  & 66.23                    & 74.95       & 74.89      \\ \hline
$\mathcal{SN}(0.1, 0.001, 0.7)$  & 66.23                    & 74.95       & 74.89      \\ \hline
$\mathcal{SN}(0.01, 0.001, 0.8)$ & 66.23                    & 74.95       & 74.89      \\ \hline
\end{tabular}
}
\end{table}




We evaluated all the priors previously exposed to understand how much the prior choice impacts the optimization problem and the final results. Only one result is shown, due to the large number of priors; the best results, for each method, are then used to train the models for all the experiments; the overall classification results will be presented later. 

\begin{figure}[!t]
\centering
  \begin{subfigure}[b]{0.35\textwidth}
  \centering
    \includegraphics[width=\textwidth]{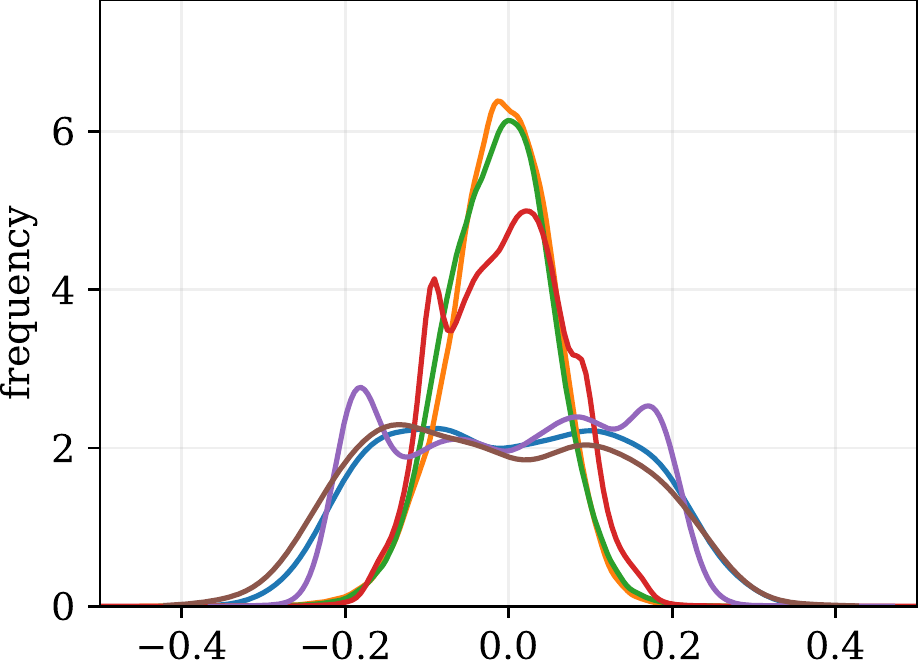}
    \caption{MMD}
  \end{subfigure}
  \begin{subfigure}[b]{0.35\textwidth}
  \centering
    \includegraphics[width=\textwidth]{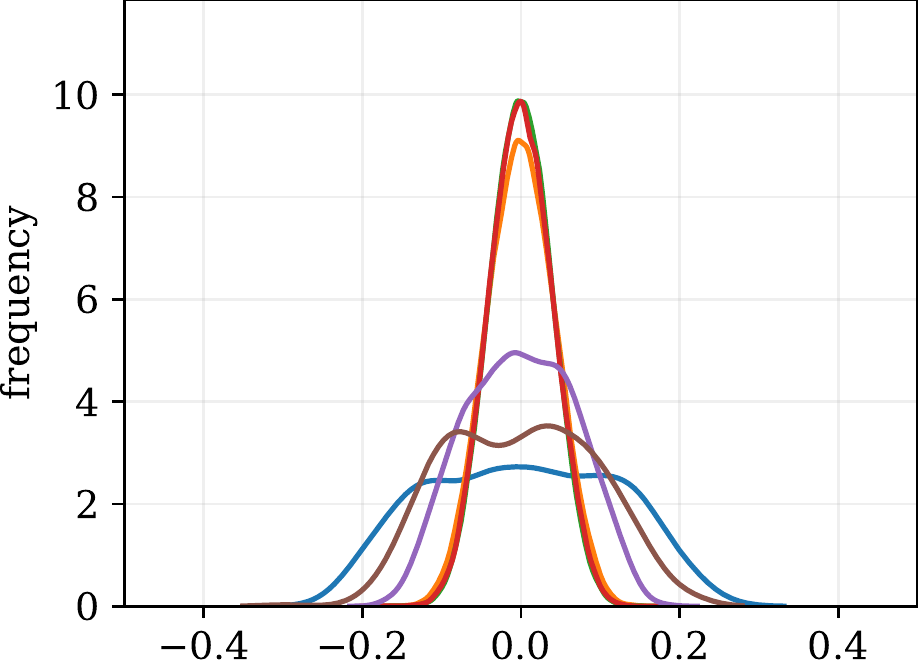}
    \caption{BBB}
  \end{subfigure}
\label{im:hist}
  \caption{The images show the posterior distribution of the weights obtained on CIFAR10 with the prior $\text{Laplace}(0, 1)$. BBB method fails when combined with the peaked prior, because it forces the convergence of the distributions on zero, neglecting the minimization problem associated to the classification. Each color represents the weights of a specific layer.}
  \label{img:distributions}
\end{figure}

The Table \ref{table:priors} shows the results obtained, on CIFAR10, with all the tested priors. It is clear that BBB fails to converge with spiky priors because the KL divergence forces the distributions to collapse on zero.
A clear case of this behaviour can be observed with the Laplacian prior, as shown in Fig. \ref{img:distributions}.

In the end, we can say that MMD works better than BBB, even with an uninformative prior, such as a uniform distribution which gives only a range for the parameters, because its sampling nature allows more operating space than BBB. Moreover, Fig. \ref{img:distributions} also shows that BNNs trained with MMD are capable of approximating a more complex posterior. 

\begin{figure*}[!t]
\centering
  \begin{subfigure}[b]{0.32\textwidth}
  \centering
    \includegraphics[width=\textwidth]{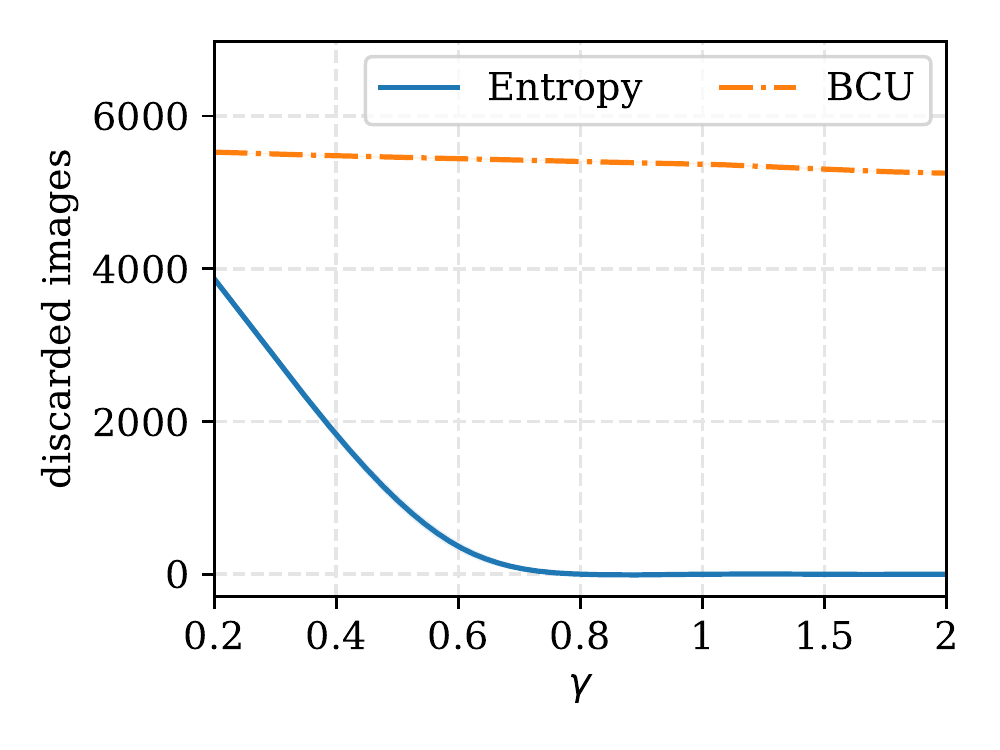}
    \caption{Discarded images while varying the threshold $\gamma$.}
  \end{subfigure}
  \hfill
  \begin{subfigure}[b]{0.32\textwidth}
  \centering
    \includegraphics[width=\textwidth]{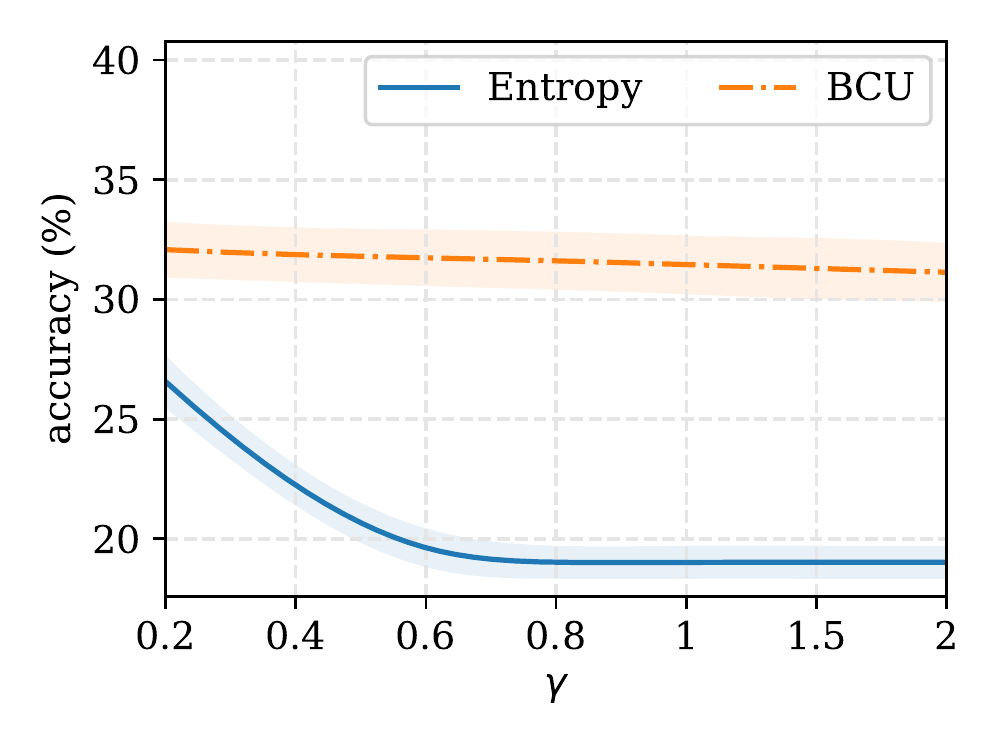}
    \caption{Classification score while varying the threshold $\gamma$.}
  \end{subfigure}   
  \hfill
  \begin{subfigure}[b]{0.32\textwidth}
   \centering
\includegraphics[width=\textwidth]{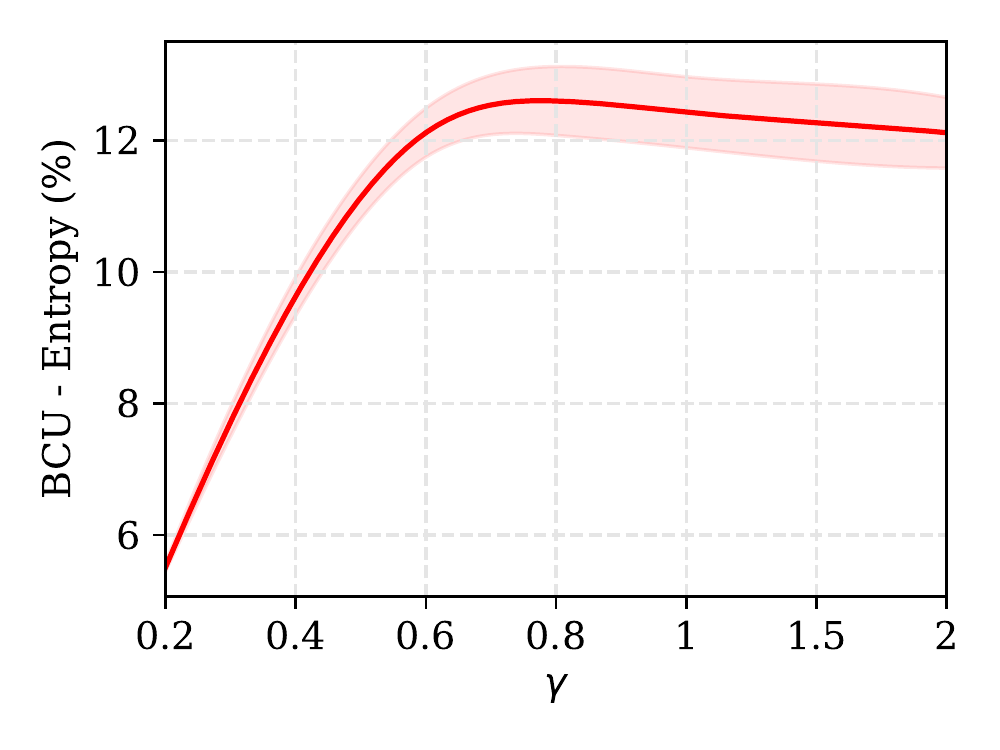}
\caption{Difference between the scores obtained.}
  \end{subfigure}
  \caption{The images show, respectively, how many images are discarded, the obtained score calculated over the samples that have not been discarded, and, in the last plot, the difference between the classification score obtained using BCU and the entropy based thresholds. We tested different $\gamma$ thresholds. The results are associated to the best model trained on CIFAR100 with the BNN trained using the proposed MMD method, under the FGSM attack with $\epsilon=0.005$.}
  \label{fig:cifar100_fgsm}
\end{figure*}

\begin{figure*}[!t]
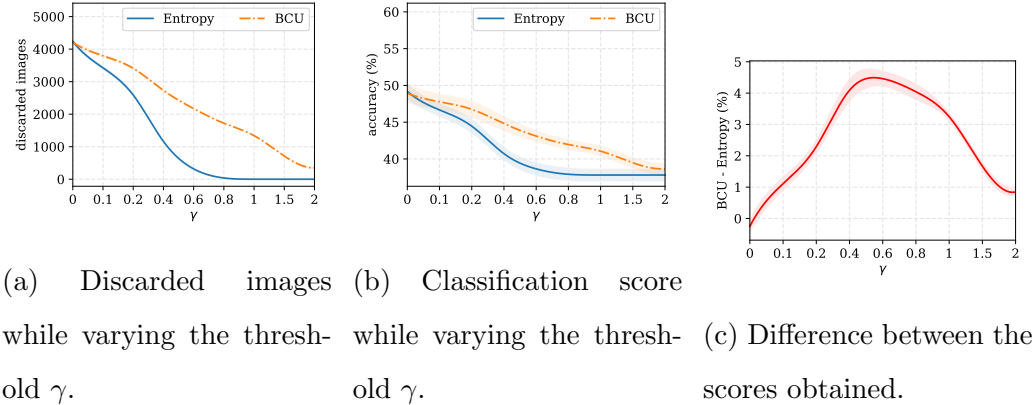

\centering
  \begin{subfigure}[b]{0.32\textwidth}
    \includegraphics[width=\textwidth]{images/unc/dropout_cifar10/{{0.001_disc}}}
    \caption{Discarded images while varying the threshold $\gamma$.}
  \end{subfigure}
  \hfill
  \begin{subfigure}[b]{0.32\textwidth}
    \includegraphics[width=\textwidth]{images/unc/dropout_cifar10/{{0.001_score}}}
    \caption{Classification score while varying the threshold $\gamma$.}
  \end{subfigure}
  \hfill
  \begin{subfigure}[b]{0.32\textwidth}
\includegraphics[width=\textwidth]{images/unc/dropout_cifar10/{{0.001_score_difference}}}
\caption{Difference between the scores obtained.}
  \end{subfigure}
  \caption{The images show, respectively, how many images are discarded, the obtained score calculated over the samples that have not been discarded, and, in the last plot, the difference between the classification score obtained using the BCU and entropy based thresholds. We tested different $\gamma$ thresholds. The results are associated to the best model trained on CIFAR10 with the MC Dropout approach, under the FGSM attack with $\epsilon=0.001$.}
  \label{fig:cifar10_fgsm}
\end{figure*}

\begin{table}[t]
\centering
\caption{The Table shows, for each method, the results with the associated standard deviation, in term of calibration, measured as ECE score (\%, Eq. \eqref{eq:ECE})); lower is better.}
\resizebox{0.7\textwidth}{!}{
\begin{tabular}{c|c|c|c|c|}
\cline{2-5}
\multicolumn{1}{l|}{}          & DNN               & MC Dropout           & \multicolumn{1}{c|}{\begin{tabular}[c]{@{}c@{}}MC Dropout\\  (no weight decay)\end{tabular}} & \begin{tabular}[c]{@{}c@{}}MMD  \\ (Weight wise)  \end{tabular} \\ \hline
\multicolumn{1}{|c|}{MNIST}    & 0.73 $\rpm$ 0.09 & \bt 0.41 $\rpm$ 0.11 & 0.49 $\rpm$ 0.23                                                                             & 0.50 $\rpm$ 0.08        \\ \hline
\multicolumn{1}{|c|}{CIFAR10}  & 14.56 $\rpm$ 0.32 & 3.43 $\rpm$ 0.57    & 6.00 $\rpm$ 0.17                                                                             & \bt 5.93  $\rpm$ 0.91 \\ \hline
\multicolumn{1}{|c|}{CIFAR100} & 13.11 $\rpm$ 6.75 & \bt 2.22 $\rpm$ 0.63 & 5.92 $\rpm$ 0.58                                                                             & 3.89 $\rpm$ 1.76       \\ \hline
\end{tabular}%
}
\label{table:ece}
\end{table}
\label{subsec:calibration}
\begin{figure*}[!t]
\centering
  \begin{subfigure}[t]{0.32\textwidth}
  \centering
    \includegraphics[width=\textwidth]{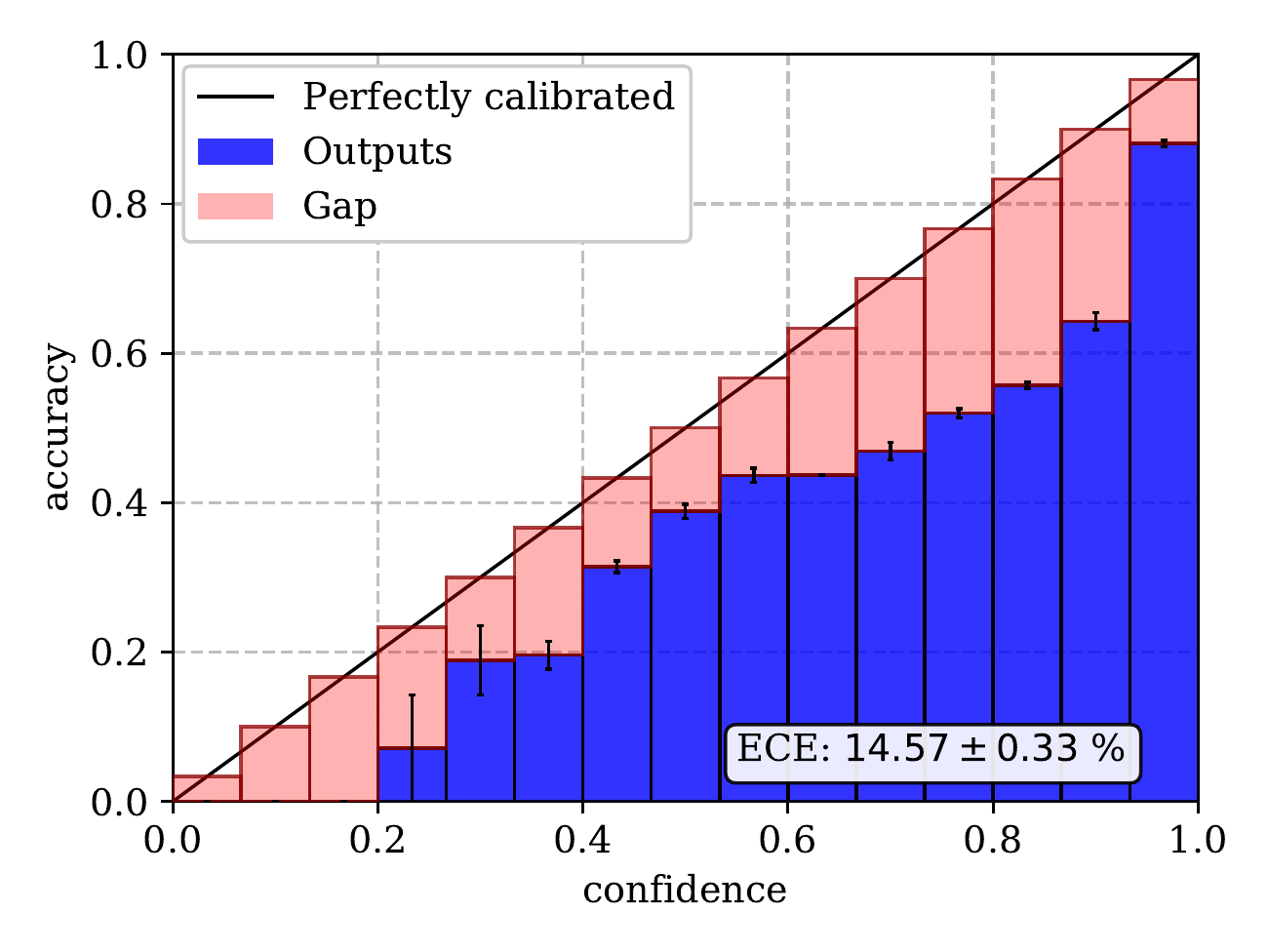}
    \caption{Reliability diagram of DNN.}
  \end{subfigure}
  \hfill
  \begin{subfigure}[t]{0.32\textwidth}
  \centering
    \includegraphics[width=\textwidth]{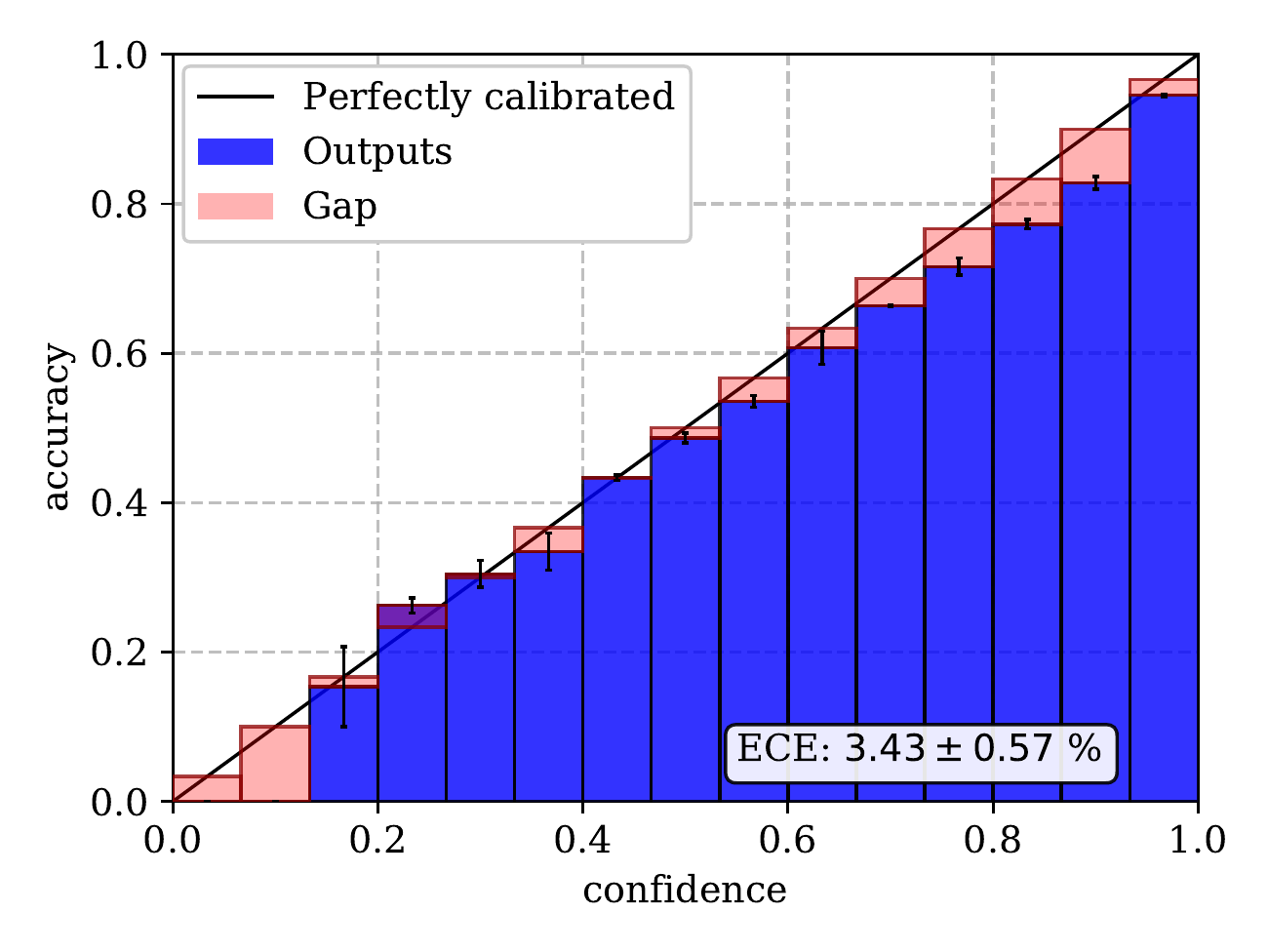}
    \caption{Reliability diagram of MC Dropout.}
  \end{subfigure}   
  \hfill
\begin{subfigure}[t]{0.32\textwidth}
\centering
\includegraphics[width=\textwidth]{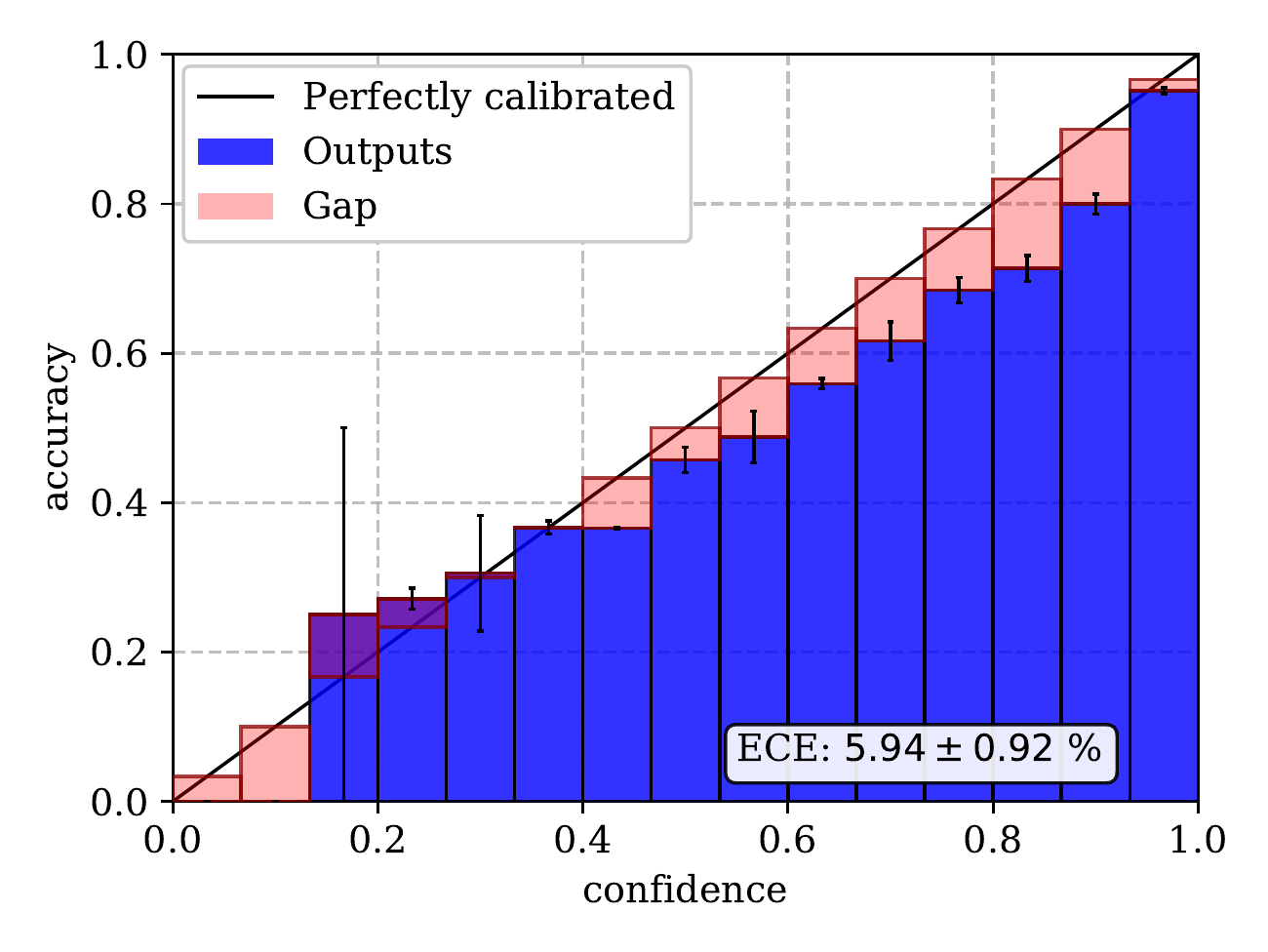}
    \caption{Reliability diagram of MMD.}
  \end{subfigure}
  \caption{The images show the reliability diagram for each method compared in Table \ref{table:ece}. In these images the correlation between the ECE score and the gap bars is shown visually. The methods are trained on CIFAR10.}
  \label{fig:rel_diagram}
\end{figure*}

\subsubsection{Classification results}

Table \ref{table:score} shows the results obtained on the classification experiments. It shows that BBB method fails drastically if we use a weight-wise posterior, but also to reach good performances when the posterior is neuron-wise and the dataset becomes harder (CIFAR100). In the end we can say that the networks trained with the original ELBO loss fail when the models become bigger and the dataset harder; we will also show that they are also more sensible to the choice of the prior. 

\subsubsection{FGSM test}
\label{subsec:fgsm_exp}

In this test we compare the proposed BCU measure \eqref{eq:var_hat} with the normalized entropy formulation in \eqref{eq:entropy}, under the FGSM attack \cite{goodfellow2014explaining}, in which, given an image $x$ and its label $y$, we modify the image $x$ as: $x + \epsilon \ \text{sign}(\nabla_x J(\theta, x, y))$, where $J$ is the input-output Jacobian of a randomly sampled network. 

The purpose is to discard images in which the network is less confident, therefore we study how the threshold, defined as in Eq. \eqref{eq:thres}, behaves when we change the uncertainty measure.

In Fig. \ref{fig:cifar10_fgsm} and \ref{fig:cifar100_fgsm}, we show the results obtained, respectively, on CIFAR10 with MC Dropout and CIFAR100 with MMD. We can see how the number of discarded images decrease exponentially when the threshold is applied to the uncertainty based entropy measure; the score also drops, since more noisy images are evaluated instead of being discarded. 
This is due to the fact that the entropy measure does not take into consideration the correlation between the classes, and this happens because only the distribution obtained using a set of weights $\mathbf{w}_t$ is evaluated at each time, thus the entropy does not codify the overall uncertainty across all the possible models and how a class can influence the others. Both of these informations are taken into account when using the measure of uncertainty proposed.

\subsubsection{Network calibration}

In this section, we evaluate the calibration of each network. To visually show the calibration of these models we used the reliability diagrams \cite{degroot1983comparison, niculescu2005predicting}. Fig. \ref{fig:rel_diagram} shows these diagrams, while Table \ref{table:score} contains the results achieved in terms of ECE score. Only the networks that achieve a classification result near their best one, presented in Table \ref{table:score}, are considered in this experiments; for this reason, results obtained with BNNs trained with BBB method are not evaluated due to the inability of reaching competitive scores. We decided also to compare two different versions of MC Dropout to make the comparisons fairer, because the original one uses a weight decay, which leads to a better ECE score (as pointed out in \cite{guo2017calibration}); consequently we trained also a MC Dropout network without weight regularization. 
We can observe that DNN never achieves a good calibration, and while MC Dropout networks are well calibrated due to the weight decay, our method achieves a good calibration result even if no regularization is used. By comparing our method with the MC Dropout without weight regularization, we find that our method achieves a better ECE score. In the end we can say that the BNNs trained using MMD, in general, are well calibrated and do not require external normalization techniques to achieve it. 

\section{Conclusion}
In this paper, we proposed a new VI method to approximate the posterior over the weights of a BNN, which uses the MMD distance as a regularization metric between the posterior and the prior. This method has advantageous  characteristics, if compared to other VI methods such as MC Dropout and BBB. First, the BNNs trained with this technique achieve better results, and they are able of approximating a more complex posterior. Second, it is more robust to the prior choice, if compared to BBB, an important aspect in these models. Third, this method, if combined with the right prior, can lead to a very well calibrated network, that also achieves good performance. 

We also proposed and tested a new method to calculate the classification's uncertainty of a BNN. We showed that this measure, combined with a threshold-based rejection technique, behaves better when discarding samples on which the BNN is less certain, by leading to a better score, if compared to the entropy measure, on noisy samples. 

Our MMD method suggests interesting lines of further research, in which a BNN network can be trained using VI methods that involve a regularization method different from the KL divergence, and leading to better and more interesting posteriors. 

\bibliographystyle{elsarticle-harv}
\bibliography{ref}

\begin{thebibliography}{38}
\expandafter\ifx\csname natexlab\endcsname\relax\def\natexlab#1{#1}\fi
\providecommand{\url}[1]{\texttt{#1}}
\providecommand{\href}[2]{#2}
\providecommand{\path}[1]{#1}
\providecommand{\DOIprefix}{doi:}
\providecommand{\ArXivprefix}{arXiv:}
\providecommand{\URLprefix}{URL: }
\providecommand{\Pubmedprefix}{pmid:}
\providecommand{\doi}[1]{\href{http://dx.doi.org/#1}{\path{#1}}}
\providecommand{\Pubmed}[1]{\href{pmid:#1}{\path{#1}}}
\providecommand{\bibinfo}[2]{#2}
\ifx\xfnm\relax \def\xfnm[#1]{\unskip,\space#1}\fi
\bibitem[{Alemi et~al.(2017)Alemi, Poole, Fischer, Dillon, Saurous and
  Murphy}]{alemi2017fixing}
\bibinfo{author}{Alemi, A.A.}, \bibinfo{author}{Poole, B.},
  \bibinfo{author}{Fischer, I.}, \bibinfo{author}{Dillon, J.V.},
  \bibinfo{author}{Saurous, R.A.}, \bibinfo{author}{Murphy, K.},
  \bibinfo{year}{2017}.
\newblock \bibinfo{title}{Fixing a broken {ELBO}}.
\newblock \bibinfo{journal}{arXiv preprint arXiv:1711.00464} .
\bibitem[{Blei et~al.(2017)Blei, Kucukelbir and
  McAuliffe}]{blei2017variational}
\bibinfo{author}{Blei, D.M.}, \bibinfo{author}{Kucukelbir, A.},
  \bibinfo{author}{McAuliffe, J.D.}, \bibinfo{year}{2017}.
\newblock \bibinfo{title}{Variational inference: A review for statisticians}.
\newblock \bibinfo{journal}{Journal of the American statistical Association}
  \bibinfo{volume}{112}, \bibinfo{pages}{859--877}.
\bibitem[{Blundell et~al.(2015)Blundell, Cornebise, Kavukcuoglu and
  Wierstra}]{blundell2015weight}
\bibinfo{author}{Blundell, C.}, \bibinfo{author}{Cornebise, J.},
  \bibinfo{author}{Kavukcuoglu, K.}, \bibinfo{author}{Wierstra, D.},
  \bibinfo{year}{2015}.
\newblock \bibinfo{title}{Weight uncertainty in neural networks}, in:
  \bibinfo{booktitle}{Proceedings of the 32nd International Conference on
  Machine Learning (ICML)}.
\bibitem[{Briol et~al.(2019)Briol, Barp, Duncan and
  Girolami}]{briol2019statistical}
\bibinfo{author}{Briol, F.X.}, \bibinfo{author}{Barp, A.},
  \bibinfo{author}{Duncan, A.B.}, \bibinfo{author}{Girolami, M.},
  \bibinfo{year}{2019}.
\newblock \bibinfo{title}{Statistical inference for generative models with
  maximum mean discrepancy}.
\newblock \bibinfo{journal}{arXiv preprint arXiv:1906.05944} .
\bibitem[{Buntine(1991)}]{buntine1991bayesian}
\bibinfo{author}{Buntine, W.L.}, \bibinfo{year}{1991}.
\newblock \bibinfo{title}{Bayesian back-propagation}.
\newblock \bibinfo{journal}{Complex Systems} \bibinfo{volume}{5},
  \bibinfo{pages}{603--643}.
\bibitem[{Chen et~al.(2015)Chen, Ding and Carin}]{chen2015convergence}
\bibinfo{author}{Chen, C.}, \bibinfo{author}{Ding, N.}, \bibinfo{author}{Carin,
  L.}, \bibinfo{year}{2015}.
\newblock \bibinfo{title}{On the convergence of stochastic gradient {MCMC}
  algorithms with high-order integrators}, in: \bibinfo{booktitle}{Advances in
  Neural Information Processing Systems}, pp. \bibinfo{pages}{2278--2286}.
\bibitem[{Ch{\'e}rief-Abdellatif and Alquier(2019)}]{cherief2019finite}
\bibinfo{author}{Ch{\'e}rief-Abdellatif, B.E.}, \bibinfo{author}{Alquier, P.},
  \bibinfo{year}{2019}.
\newblock \bibinfo{title}{Finite sample properties of parametric mmd
  estimation: robustness to misspecification and dependence}.
\newblock \bibinfo{journal}{arXiv preprint arXiv:1912.05737} .
\bibitem[{Cherief-Abdellatif and Alquier(2020)}]{abdellatif20a}
\bibinfo{author}{Cherief-Abdellatif, B.E.}, \bibinfo{author}{Alquier, P.},
  \bibinfo{year}{2020}.
\newblock \bibinfo{title}{Mmd-bayes: Robust bayesian estimation via maximum
  mean discrepancy}, \bibinfo{publisher}{PMLR}. pp. \bibinfo{pages}{1--21}.
\newblock \URLprefix
  \url{http://proceedings.mlr.press/v118/cherief-abdellatif20a.html}.
\bibitem[{DeGroot and Fienberg(1983)}]{degroot1983comparison}
\bibinfo{author}{DeGroot, M.H.}, \bibinfo{author}{Fienberg, S.E.},
  \bibinfo{year}{1983}.
\newblock \bibinfo{title}{The comparison and evaluation of forecasters}.
\newblock \bibinfo{journal}{Journal of the Royal Statistical Society: Series D
  (The Statistician)} \bibinfo{volume}{32}, \bibinfo{pages}{12--22}.
\bibitem[{Der~Kiureghian and Ditlevsen(2009)}]{der2009aleatory}
\bibinfo{author}{Der~Kiureghian, A.}, \bibinfo{author}{Ditlevsen, O.},
  \bibinfo{year}{2009}.
\newblock \bibinfo{title}{Aleatory or epistemic? does it matter?}
\newblock \bibinfo{journal}{Structural Safety} \bibinfo{volume}{31},
  \bibinfo{pages}{105--112}.
\bibitem[{Gal and Ghahramani(2015a)}]{gal2015bayesian}
\bibinfo{author}{Gal, Y.}, \bibinfo{author}{Ghahramani, Z.},
  \bibinfo{year}{2015}a.
\newblock \bibinfo{title}{Bayesian convolutional neural networks with
  {B}ernoulli approximate variational inference}.
\newblock \bibinfo{journal}{arXiv preprint arXiv:1506.02158} .
\bibitem[{Gal and Ghahramani(2015b)}]{gal2015dropout}
\bibinfo{author}{Gal, Y.}, \bibinfo{author}{Ghahramani, Z.},
  \bibinfo{year}{2015}b.
\newblock \bibinfo{title}{Dropout as a {B}ayesian approximation: Representing
  model uncertainty in deep learning}.
\newblock \bibinfo{journal}{arXiv preprint arXiv:1506.02142} .
\bibitem[{Goodfellow et~al.(2014)Goodfellow, Shlens and
  Szegedy}]{goodfellow2014explaining}
\bibinfo{author}{Goodfellow, I.J.}, \bibinfo{author}{Shlens, J.},
  \bibinfo{author}{Szegedy, C.}, \bibinfo{year}{2014}.
\newblock \bibinfo{title}{Explaining and harnessing adversarial examples}.
\newblock \bibinfo{journal}{arXiv preprint arXiv:1412.6572} .
\bibitem[{Graves(2011)}]{graves2011practical}
\bibinfo{author}{Graves, A.}, \bibinfo{year}{2011}.
\newblock \bibinfo{title}{Practical variational inference for neural networks},
  in: \bibinfo{booktitle}{Advances in Neural Information Processing Systems},
  pp. \bibinfo{pages}{2348--2356}.
\bibitem[{Gretton et~al.(2012)Gretton, Borgwardt, Rasch, Sch\"{o}lkopf and
  Smola}]{Gretton:2012:KTT:2188385.2188410}
\bibinfo{author}{Gretton, A.}, \bibinfo{author}{Borgwardt, K.M.},
  \bibinfo{author}{Rasch, M.J.}, \bibinfo{author}{Sch\"{o}lkopf, B.},
  \bibinfo{author}{Smola, A.}, \bibinfo{year}{2012}.
\newblock \bibinfo{title}{A kernel two-sample test}.
\newblock \bibinfo{journal}{The Journal of Machine Learning Research}
  \bibinfo{volume}{13}, \bibinfo{pages}{723--773}.
\bibitem[{Guo et~al.(2017)Guo, Pleiss, Sun and Weinberger}]{guo2017calibration}
\bibinfo{author}{Guo, C.}, \bibinfo{author}{Pleiss, G.}, \bibinfo{author}{Sun,
  Y.}, \bibinfo{author}{Weinberger, K.Q.}, \bibinfo{year}{2017}.
\newblock \bibinfo{title}{On calibration of modern neural networks}, in:
  \bibinfo{booktitle}{Proceedings of the 34th International Conference on
  Machine Learning-Volume 70}, \bibinfo{organization}{JMLR. org}. pp.
  \bibinfo{pages}{1321--1330}.
\bibitem[{He et~al.(2015)He, Zhang, Ren and Sun}]{he2015delving}
\bibinfo{author}{He, K.}, \bibinfo{author}{Zhang, X.}, \bibinfo{author}{Ren,
  S.}, \bibinfo{author}{Sun, J.}, \bibinfo{year}{2015}.
\newblock \bibinfo{title}{Delving deep into rectifiers: Surpassing human-level
  performance on {I}magenet classification}, in:
  \bibinfo{booktitle}{Proceedings of the IEEE International Conference on
  Computer Vision}, pp. \bibinfo{pages}{1026--1034}.
\bibitem[{Hinton and Van~Camp(1993)}]{hinton1993keeping}
\bibinfo{author}{Hinton, G.E.}, \bibinfo{author}{Van~Camp, D.},
  \bibinfo{year}{1993}.
\newblock \bibinfo{title}{Keeping the neural networks simple by minimizing the
  description length of the weights}, in: \bibinfo{booktitle}{Proceedings of
  the sixth annual conference on Computational learning theory}, pp.
  \bibinfo{pages}{5--13}.
\bibitem[{H{\"u}llermeier and Waegeman(2019)}]{hullermeier2019aleatoric}
\bibinfo{author}{H{\"u}llermeier, E.}, \bibinfo{author}{Waegeman, W.},
  \bibinfo{year}{2019}.
\newblock \bibinfo{title}{Aleatoric and epistemic uncertainty in machine
  learning: A tutorial introduction}.
\newblock \bibinfo{journal}{arXiv preprint arXiv:1910.09457} .
\bibitem[{{Karolina Dziugaite} et~al.(2015){Karolina Dziugaite}, {Roy} and
  {Ghahramani}}]{2015arXiv150503906K}
\bibinfo{author}{{Karolina Dziugaite}, G.}, \bibinfo{author}{{Roy}, D.M.},
  \bibinfo{author}{{Ghahramani}, Z.}, \bibinfo{year}{2015}.
\newblock \bibinfo{title}{{Training generative neural networks via Maximum Mean
  Discrepancy optimization}}.
\newblock \bibinfo{journal}{arXiv e-prints} ,
  \bibinfo{pages}{arXiv:1505.03906}\href{http://arxiv.org/abs/1505.03906}{{\tt
  arXiv:1505.03906}}.
\bibitem[{Kendall and Gal(2017)}]{kendall2017uncertainties}
\bibinfo{author}{Kendall, A.}, \bibinfo{author}{Gal, Y.}, \bibinfo{year}{2017}.
\newblock \bibinfo{title}{What uncertainties do we need in {B}ayesian deep
  learning for computer vision?}, in: \bibinfo{booktitle}{Advances in Neural
  Information Processing Systems}, pp. \bibinfo{pages}{5574--5584}.
\bibitem[{Kingma and Ba(2014)}]{kingma2014adam}
\bibinfo{author}{Kingma, D.P.}, \bibinfo{author}{Ba, J.}, \bibinfo{year}{2014}.
\newblock \bibinfo{title}{Adam: A method for stochastic optimization}.
\newblock \bibinfo{journal}{arXiv preprint arXiv:1412.6980} .
\bibitem[{Kingma et~al.(2015)Kingma, Salimans and
  Welling}]{kingma2015variational}
\bibinfo{author}{Kingma, D.P.}, \bibinfo{author}{Salimans, T.},
  \bibinfo{author}{Welling, M.}, \bibinfo{year}{2015}.
\newblock \bibinfo{title}{Variational dropout and the local reparameterization
  trick}, in: \bibinfo{booktitle}{Advances in Neural Information Processing
  Systems}, pp. \bibinfo{pages}{2575--2583}.
\bibitem[{Kullback and Leibler(1951)}]{kullback1951information}
\bibinfo{author}{Kullback, S.}, \bibinfo{author}{Leibler, R.A.},
  \bibinfo{year}{1951}.
\newblock \bibinfo{title}{On information and sufficiency}.
\newblock \bibinfo{journal}{The Annals of Mathematical Statistics}
  \bibinfo{volume}{22}, \bibinfo{pages}{79--86}.
\bibitem[{Kwon et~al.(2020)Kwon, Won, Kim and Paik}]{kwon2020uncertainty}
\bibinfo{author}{Kwon, Y.}, \bibinfo{author}{Won, J.H.}, \bibinfo{author}{Kim,
  B.J.}, \bibinfo{author}{Paik, M.C.}, \bibinfo{year}{2020}.
\newblock \bibinfo{title}{Uncertainty quantification using {B}ayesian neural
  networks in classification: Application to biomedical image segmentation}.
\newblock \bibinfo{journal}{Computational Statistics \& Data Analysis}
  \bibinfo{volume}{142}, \bibinfo{pages}{106816}.
\bibitem[{LeCun et~al.(2010)LeCun, Cortes and Burges}]{lecun2010mnist}
\bibinfo{author}{LeCun, Y.}, \bibinfo{author}{Cortes, C.},
  \bibinfo{author}{Burges, C.}, \bibinfo{year}{2010}.
\newblock \bibinfo{title}{Mnist handwritten digit database}.
\newblock \bibinfo{journal}{ATT Labs [Online]. Available: http://yann. lecun.
  com/exdb/mnist} \bibinfo{volume}{2}.
\bibitem[{Leibig et~al.(2017)Leibig, Allken, Ayhan, Berens and
  Wahl}]{leibig2017leveraging}
\bibinfo{author}{Leibig, C.}, \bibinfo{author}{Allken, V.},
  \bibinfo{author}{Ayhan, M.S.}, \bibinfo{author}{Berens, P.},
  \bibinfo{author}{Wahl, S.}, \bibinfo{year}{2017}.
\newblock \bibinfo{title}{Leveraging uncertainty information from deep neural
  networks for disease detection}.
\newblock \bibinfo{journal}{Scientific Reports} \bibinfo{volume}{7},
  \bibinfo{pages}{1--14}.
\bibitem[{Li et~al.(2017)Li, Chang, Cheng, Yang and P{\'o}czos}]{li2017mmd}
\bibinfo{author}{Li, C.L.}, \bibinfo{author}{Chang, W.C.},
  \bibinfo{author}{Cheng, Y.}, \bibinfo{author}{Yang, Y.},
  \bibinfo{author}{P{\'o}czos, B.}, \bibinfo{year}{2017}.
\newblock \bibinfo{title}{{MMD GAN}: Towards deeper understanding of moment
  matching network}, in: \bibinfo{booktitle}{Advances in Neural Information
  Processing Systems}, pp. \bibinfo{pages}{2203--2213}.
\bibitem[{Li et~al.(2015)Li, Swersky and Zemel}]{DBLP:journals/corr/LiSZ15}
\bibinfo{author}{Li, Y.}, \bibinfo{author}{Swersky, K.},
  \bibinfo{author}{Zemel, R.S.}, \bibinfo{year}{2015}.
\newblock \bibinfo{title}{Generative moment matching networks}.
\newblock \bibinfo{journal}{arXiv:1502.02761} .
\bibitem[{MacKay(1992)}]{mackay1992practical}
\bibinfo{author}{MacKay, D.J.}, \bibinfo{year}{1992}.
\newblock \bibinfo{title}{A practical {B}ayesian framework for backpropagation
  networks}.
\newblock \bibinfo{journal}{Neural Computation} \bibinfo{volume}{4},
  \bibinfo{pages}{448--472}.
\bibitem[{MacKay and Mac~Kay(2003)}]{mackay2003information}
\bibinfo{author}{MacKay, D.J.}, \bibinfo{author}{Mac~Kay, D.J.},
  \bibinfo{year}{2003}.
\newblock \bibinfo{title}{Information theory, inference and learning
  algorithms}.
\newblock \bibinfo{publisher}{Cambridge University Press}.
\bibitem[{Naeini et~al.(2015)Naeini, Cooper and
  Hauskrecht}]{naeini2015obtaining}
\bibinfo{author}{Naeini, M.P.}, \bibinfo{author}{Cooper, G.},
  \bibinfo{author}{Hauskrecht, M.}, \bibinfo{year}{2015}.
\newblock \bibinfo{title}{Obtaining well calibrated probabilities using
  bayesian binning}, in: \bibinfo{booktitle}{Twenty-Ninth AAAI Conference on
  Artificial Intelligence}.
\bibitem[{Neal(1996)}]{neal2012bayesian}
\bibinfo{author}{Neal, R.M.}, \bibinfo{year}{1996}.
\newblock \bibinfo{title}{Bayesian Learning for Neural Networks}.
\newblock \bibinfo{publisher}{Springer-Verlag}, \bibinfo{address}{Berlin,
  Heidelberg}.
\bibitem[{Niculescu-Mizil and Caruana(2005)}]{niculescu2005predicting}
\bibinfo{author}{Niculescu-Mizil, A.}, \bibinfo{author}{Caruana, R.},
  \bibinfo{year}{2005}.
\newblock \bibinfo{title}{Predicting good probabilities with supervised
  learning}, in: \bibinfo{booktitle}{Proceedings of the 22nd International
  Conference on Machine Learning}, \bibinfo{publisher}{Association for
  Computing Machinery}, \bibinfo{address}{New York, NY, USA}. p.
  \bibinfo{pages}{625–632}.
\newblock \URLprefix \url{https://doi.org/10.1145/1102351.1102430},
  \DOIprefix\doi{10.1145/1102351.1102430}.
\bibitem[{Rethage et~al.(2018)Rethage, Pons and Serra}]{rethage2018wavenet}
\bibinfo{author}{Rethage, D.}, \bibinfo{author}{Pons, J.},
  \bibinfo{author}{Serra, X.}, \bibinfo{year}{2018}.
\newblock \bibinfo{title}{A wavenet for speech denoising}, in:
  \bibinfo{booktitle}{2018 IEEE International Conference on Acoustics, Speech
  and Signal Processing (ICASSP)}, \bibinfo{organization}{IEEE}. pp.
  \bibinfo{pages}{5069--5073}.
\bibitem[{{Shridhar} et~al.(2019){Shridhar}, {Laumann} and
  {Liwicki}}]{2019arXiv190102731S}
\bibinfo{author}{{Shridhar}, K.}, \bibinfo{author}{{Laumann}, F.},
  \bibinfo{author}{{Liwicki}, M.}, \bibinfo{year}{2019}.
\newblock \bibinfo{title}{{A Comprehensive guide to Bayesian Convolutional
  Neural Network with Variational Inference}}.
\newblock \bibinfo{journal}{arXiv e-prints} ,
  \bibinfo{pages}{arXiv:1901.02731}\href{http://arxiv.org/abs/1901.02731}{{\tt
  arXiv:1901.02731}}.
\bibitem[{Wenzel et~al.(2020)Wenzel, Roth, Veeling, Światkowski, Tran, Mandt,
  Snoek, Salimans, Jenatton and Nowozin}]{wenzel2020good}
\bibinfo{author}{Wenzel, F.}, \bibinfo{author}{Roth, K.},
  \bibinfo{author}{Veeling, B.S.}, \bibinfo{author}{Światkowski, J.},
  \bibinfo{author}{Tran, L.}, \bibinfo{author}{Mandt, S.},
  \bibinfo{author}{Snoek, J.}, \bibinfo{author}{Salimans, T.},
  \bibinfo{author}{Jenatton, R.}, \bibinfo{author}{Nowozin, S.},
  \bibinfo{year}{2020}.
\newblock \bibinfo{title}{How good is the {B}ayes posterior in deep neural
  networks really?}
\newblock \bibinfo{journal}{arXiv:2002.02405} .
\bibitem[{{Zhao} et~al.(2017){Zhao}, {Song} and {Ermon}}]{2017arXiv170602262Z}
\bibinfo{author}{{Zhao}, S.}, \bibinfo{author}{{Song}, J.},
  \bibinfo{author}{{Ermon}, S.}, \bibinfo{year}{2017}.
\newblock \bibinfo{title}{{InfoVAE: Information Maximizing Variational
  Autoencoders}}.
\newblock \bibinfo{journal}{arXiv e-prints} ,
  \bibinfo{pages}{arXiv:1706.02262}\href{http://arxiv.org/abs/1706.02262}{{\tt
  arXiv:1706.02262}}.

\end{thebibliography}


\end{document}